\documentclass[12pt]{iopart}

\usepackage{bm}             
\usepackage{amsfonts}       
\expandafter\let\csname equation*\endcsname\relax
\expandafter\let\csname endequation*\endcsname\relax
\usepackage{amsmath}       
\usepackage{hyperref}
\usepackage{graphicx}
\usepackage[usenames,dvipsnames]{xcolor}
\usepackage{enumitem}
\usepackage{physics}
\usepackage[numbers,sort&compress]{natbib}
\usepackage{algorithm}
\usepackage{algpseudocode}

\newcommand*{\tran}{^{\mkern-1.5mu\mathsf{T}}}
\newcommand\eqdef{\mathrel{\overset{\makebox[0pt]{\mbox{\normalfont\tiny\sffamily $\Delta$}}}{=}}}

\hypersetup{
    colorlinks=true,
    citecolor=Blue,
    linkcolor=Mahogany,
    urlcolor=teal
}

\begin{document}

\title[\footnotesize The Role of the Time-Dependent Hessian in High-Dimensional Optimization]{The Role of the Time-Dependent Hessian in High-Dimensional Optimization}

\author{Tony Bonnaire$^1$, Giulio Biroli$^1$, Chiara Cammarota$^3$}
\address{$^1$ Laboratoire de Physique de l'Ecole Normale Sup\'erieure, ENS, Universit\'e PSL, CNRS, Sorbonne Universit\'e, Universit\'e Paris Cit\'e, F-75005 Paris, France.}
\address{$^2$ Dipartimento di Fisica, Sapienza Università di Roma and Istituto Nazionale di Fisica Nucleare, Sezione di Roma I, P. le A. Moro 5, 00185 Rome, Italy.}

\ead{tony.bonnaire@phys.ens.fr}

\begin{abstract}
Gradient descent is commonly used to find minima in rough landscapes, particularly in recent machine learning applications. However, a theoretical understanding of why good solutions are found remains elusive, especially in strongly non-convex and high-dimensional settings. Here, we focus on the phase retrieval problem as a typical example, which has received a lot of attention recently in theoretical machine learning. 
We analyze the Hessian during gradient descent, identify a dynamical transition in its spectral properties, and relate it to the ability of escaping rough regions in the loss landscape. When the signal-to-noise ratio (SNR) is large enough, an informative negative direction exists in the Hessian at the beginning of the descent, i.e in the initial condition. While descending, a BBP transition in the spectrum takes place in finite time: the direction is lost, and the dynamics is trapped in a rugged region filled with marginally stable bad minima. Surprisingly, for finite system sizes, this window of negative curvature allows the system to recover the signal well before the theoretical SNR found for infinite sizes, emphasizing the central role of initialization and early-time dynamics for efficiently navigating rough landscapes.
\end{abstract}

\noindent{\it Keywords}: Machine Learning, Phase Retrieval, Statistical Physics, Non-convex Optimization.

\section{Introduction}

Navigating rough, non-convex, and high-dimensional energy landscapes is a central topic common to various scientific fields ranging from physics and biology to statistics and machine learning \cite{Fyodorov2004, Rico2007, Auffinger2010, Baity-Jesi2019}. Often, the goal is to find some peculiar configurations of a system linked with hidden structures in the data. These configurations are typically associated with specific minima in the landscape that one seeks to locate. For instance, this is the case in models involving planted signals in the teacher-student framework \cite{Gardner1988, Seung1992, Krzakala2009_hiding, Zdeborova2018}. Although for some problems there exists dedicated optimization procedure, the workhorse techniques to find such minima are local iterative procedures like gradient descent or its stochastic variants starting from a random configuration. Understanding why -- and to what extent -- these procedures are able to efficiently navigate complex and rugged landscapes to find meaningful solutions remains an open challenge.
These optimization techniques can be seen as physical dynamics of a system quenched to low temperature -- a problem that was intensively studied in the physics literature. Therefore it is no surprise that many recent studies (several being physics-based) have addressed this question \cite{Neyshabur2017, Belkin2018, ma2018power, Venturi2019, Mannelli2020a, Martin2024, Annesi2024}, especially in light of the remarkable success of deep learning, which heavily relies on gradient descent methods to optimize strongly non-convex loss landscapes.
A key insight from previous works \cite{Soudry2016, Cai2022} is that spurious local minima are not present in certain regimes of parameters, in particular when the signal-to-noise ratio (SNR) is large enough. As a consequence, and despite their non-convexity, landscapes become easy to descend. This suggests an explanation of the success of simple dynamics based on the ``trivialization'' of the energy landscape \cite{Fyodorov2004}, and the absence of bad minima. However, this cannot be the end of the story as it is known that bad minima are still present when optimization succeeds \cite{Baity-Jesi2019, liu2020bad}, especially when the SNR is lower leading to a mostly rough landscape with numerous irrelevant local minima \cite{Ros2019}. The challenge then becomes avoiding being trapped in suboptimal minima having a poor alignment with the underlying structure of the data.
Theoretically, the study of gradient descent for matrix-tensor PCA \cite{Mannelli2019}, and later phase retrieval \cite{Mannelli2020b}, offered a possible explanation. It showed that despite the presence of an exponential number (in the dimension) of bad minima, the dynamics can avoid them with probability one. The mechanism is related to the complexity of the loss landscape: what matters is when the bad minima with the largest basins of attraction become unstable towards the good ones, not when \emph{all} the bad ones disappear. This ``blessing'' of dimension is due to the fact that the largest basins of attraction contain the initial conditions with probability one (up to corrections which are exponentially small in the dimension).

The present work studies the interactions between the optimization and the local curvature depicted by the Hessian during the descent. Following \cite{Mannelli2020b}, we focus on phase retrieval as a model for high-dimensional landscape, and on gradient flow as optimization dynamics. We characterize the evolution of the spectral properties of the Hessian during the dynamics, and show the emergence of a new phenomenon: a dynamical \emph{Baik-Ben Arous-Pêché} \cite[BBP,][]{Baik2005} transition which takes place in the spectrum of the Hessian while the system is descending the landscape. We shall show that such a transition is crucial to characterize the gradient descent dynamics in finite dimensions.

\subsection{Settings: phase retrieval and teacher-student}

Phase retrieval aims to recover a \textit{signal}, $\bm{w}^{*} \in \mathbb{R}^N$, from the observation of $M$ absolute projections of sensing vectors $\bm{x}_i \in \mathbb{R}^N$ over it, $\{|y_i|\}_{i=1}^M$, with $y_i = \bm{x}_i\tran \bm{w}^{*}$. We consider the sensing vectors $\{\bm{x}_i\}_{i=1}^M$ as i.i.d. Gaussian with zero mean and unit norm, and the signal is drawn on the $N$-sphere with $\lVert \bm{w}^\star \rVert_2 = \sqrt{N}$.
Despite its simplistic formulation, this problem appears in various scientific fields ranging from quantum chromodynamics to astrophysics \cite{Millane1990, Harrison1993, Miao2008, Shechtman2014, Fienup2019, Wong2021} and is known to be NP-hard in general \cite{Pardalos1991}. This complexity led researchers to develop numerous algorithms relying on diverse approaches over the previous decade \cite{Candes2015, Netrapalli2015, Waldspurger2015, Chen2017a, Zhang2017, Wang2017, Wang2017a, Zhang2018}. A natural way of estimating a candidate vector $\hat{\bm{w}}$ in the absence of any prior information is to specify a loss function $\ell(y_i, \hat{y}_i)$ and optimize it iteratively through a gradient descent procedure starting from a random location in the parameter space, namely
\begin{equation}
    \hat{\bm{w}}^{(t+1)} = \hat{\bm{w}}^{(t)} - \eta \nabla \mathcal{L}(\hat{\bm{w}}^{(t)}) + \eta \mu^{(t)} \hat{\bm{w}}^{(t)}, \label{eq:dynamics_GD}
\end{equation}
where $\mathcal{L}(\hat{\bm{w}}^{(t)}) = \frac{1}{2} \sum_{i=1}^M \ell(y_i, \hat{y}_i)$, $\eta$ is a fixed learning rate, $\hat{y}_i = \bm{x}_i\tran \hat{\bm{w}}^{(t)}$ is the $i$\textsuperscript{th} estimated label and $\mu^{(t)} = \hat{\bm{w}}^{(t)} \cdot \nabla \mathcal{L}(\hat{\bm{w}}^{(t)}) / N$ encodes the spherical constraint at each time step. All gradients are evaluated with respect to ${\hat{\bm{w}}^{(t)}}$. Unless otherwise specified, the initial state is a random Gaussian vector, $\hat{\bm{w}}^{(0)} \sim \mathcal{N}(\bm{0}_N, \bm{I}_N)$.

Our analysis is performed in the teacher-student setup. One network, the \textit{teacher}, generates a set of $M$ measurements $\{y_i\}_{i=1}^M$ using a signal $\bm{w}^\star \sim \mathcal{N}(\bm{0}_N, \bm{I}_N)$. A second network with the same architecture, the \textit{student}, exploits these measurements to estimate $\bm{w}^\star$ based on the procedure described by \eqref{eq:dynamics_GD}. We are interested in the generalization ability of the student as measured by the \emph{magnetization}
\begin{equation} \label{eq:magnetization}
    m(t) = \frac{\hat{\bm{w}}^{(t)}\cdot \bm{w}^\star}{N},
\end{equation}
taking value $\pm 1$ when it produces an estimate $\hat{\bm{w}}^{(t)}$ generalizing perfectly to new samples (up to a global sign). In this paper, we call \emph{equator} the set of states $\hat{\bm{w}}^{(t)}$ having a magnetization of zero, containing for instance the initial random states when $N\to\infty$.
There are various forms of loss functions studied in the literature. In order to avoid pathologies due to rare very large values of $y_i$, we focus on a normalized version of the intensity loss function defined as
\begin{equation} \label{eq:loss}
    \ell_a(y_i, \hat{y}_i) = \frac{\left( y_i^2 - \hat{y}_i^2 \right)^2}{a + y_i^2}.
\end{equation}
The role played by the normalization is important for the conditioning of Hessian eigenspectrum, in particular ensuring the existence of a hard left edge, a crucial element of our theoretical analysis. Although the precise values at which the transitions occur may vary with the choice of the loss function, we expect the physical mechanisms at hand and the interpretation we propose in this paper to generalize well to other loss functions. While the main text focuses on $a=0.01$, we provide evidence in \ref{appendix:loss} by varying $a$.
The teacher-student setting that we study is a particular case of learning a single-index model \cite{BenArous2021, BenArous2022, Bietti2022, Arnaboldi2023, Bruna2023} in which we assume the activation function of the teacher to be known to the student. These models received much attention these past years, essentially to understand the dynamics of (online) stochastic gradient descent in the loss landscape.

\subsection{Prior analyses of phase retrieval}

Previous works showed that no estimator is able to achieve a generalization error better than a random guess for phase retrieval when $\alpha = M/N <\alpha_\mathrm{WR}=0.5$. On the other hand, perfect recovery is achievable with the approximate message passing algorithm for $\alpha>1.13$ \cite{Barbier2019}.
Many of the popular optimization methods developed over the past years rely on a careful initialization followed by an iterative algorithm in a form similar to \eqref{eq:dynamics_GD}. Such an initial guess is often provided by the leading eigenvector of a matrix function of the input data. This setup, and the transition associated to the spectral initialization in the high-dimensional limit $M, N\to \infty$ with $\alpha = M/N$ of order one, was studied in detail by several seminal previous works \cite{Barbier2019, Mondelli2019, Luo2018}.
In particular, \cite{Mondelli2019, Luo2018} identify the optimal pre-processing matrix producing a non-zero overlap between its leading eigenvector and the signal when the sensing vectors are Gaussian. These results were later extended to the more generic unitary and orthogonal case in \cite{maillard2020phase, maillard2022construction}, in which the optimal pre-processing is linked to a transition in the Hessian spectrum of the free energy landscape. In parallel, several works have thoroughly investigated whether it is possible to retrieve the signal efficiently based on a random initialization. When the entries are i.i.d. Gaussian, a number of $O(N \log^3 N)$ samples trivializes the landscape making all minima become global \cite{Sun2018}, hence enabling traditional iterative methods to find a solution independently of the initialization. This threshold was later reduced to $O(N)$ in \cite{Li2020, Cai2021, Cai2022} by adapting the form of the loss function, reducing the gap with the information-theoretic threshold of $M=N$. 

Resorting to analogies with glassy dynamics of disordered systems, \cite{Mannelli2020b} argue that the convergence of gradient descent is related to the trivialization of only a subset of bad minima. The dynamics is first trapped into peculiar high-energy bad minima, commonly called \emph{threshold states} in the physics literature. When $\alpha$ is large enough, these  states develop a negative direction and a second descent phase occurs throughout a locally convex basin until a global, well-generalizing, minimum is reached. The transition between the two phases is governed by an eigenvalue popping out of the continuous bulk of the otherwise-marginal Hessian spectrum, a phenomenon dubbed BBP transition.
A similar phenomenon arises in several random matrix problems ranging from physics and ecology to finance and computer science \cite{Saade2014, Bun2017, Mannelli2019, SaraoMannelli2020, fraboul2023artificial}.

\begin{figure*}
    \centering
    \includegraphics[width=.55\linewidth]{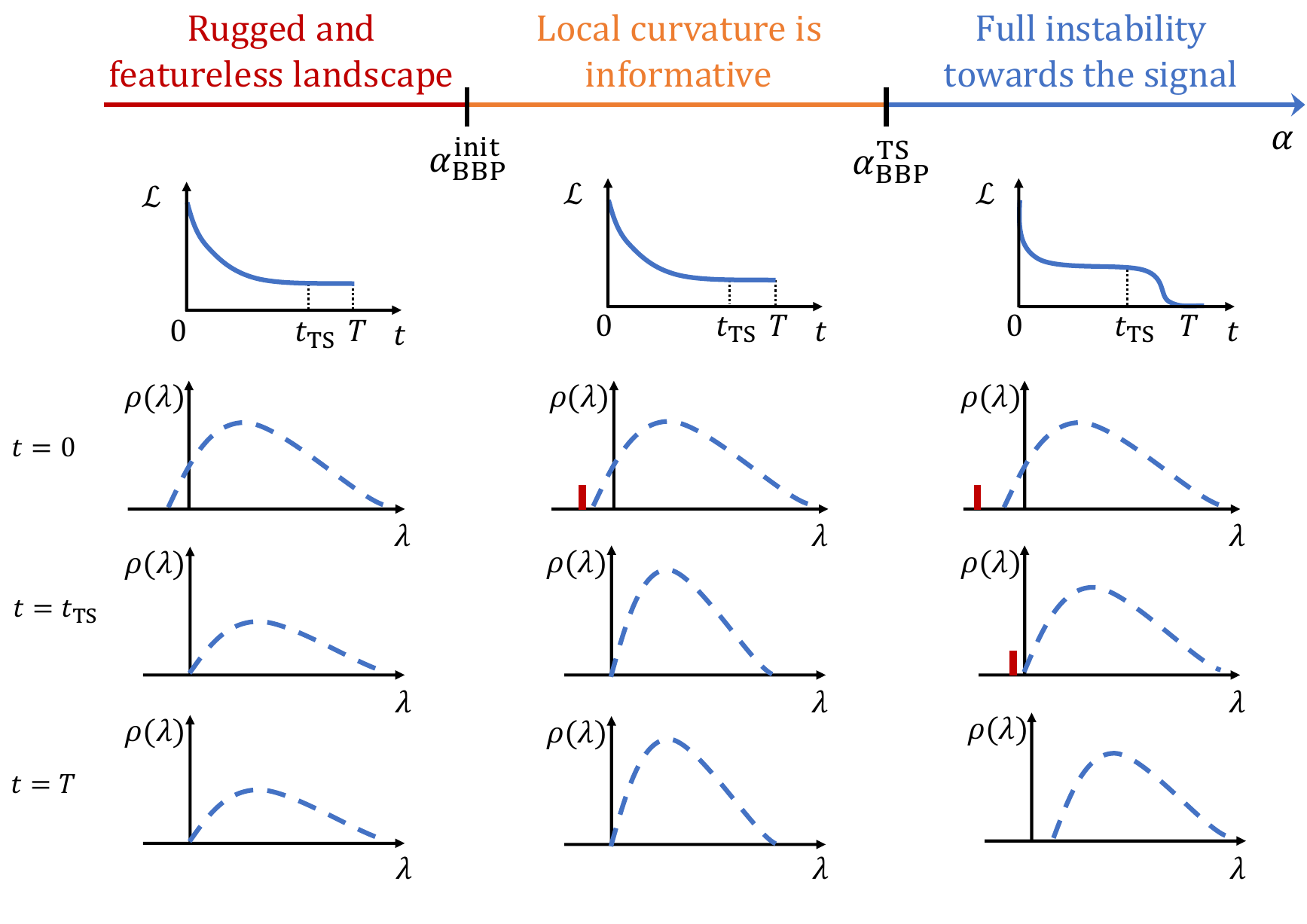}
    \includegraphics[width=.34\linewidth]{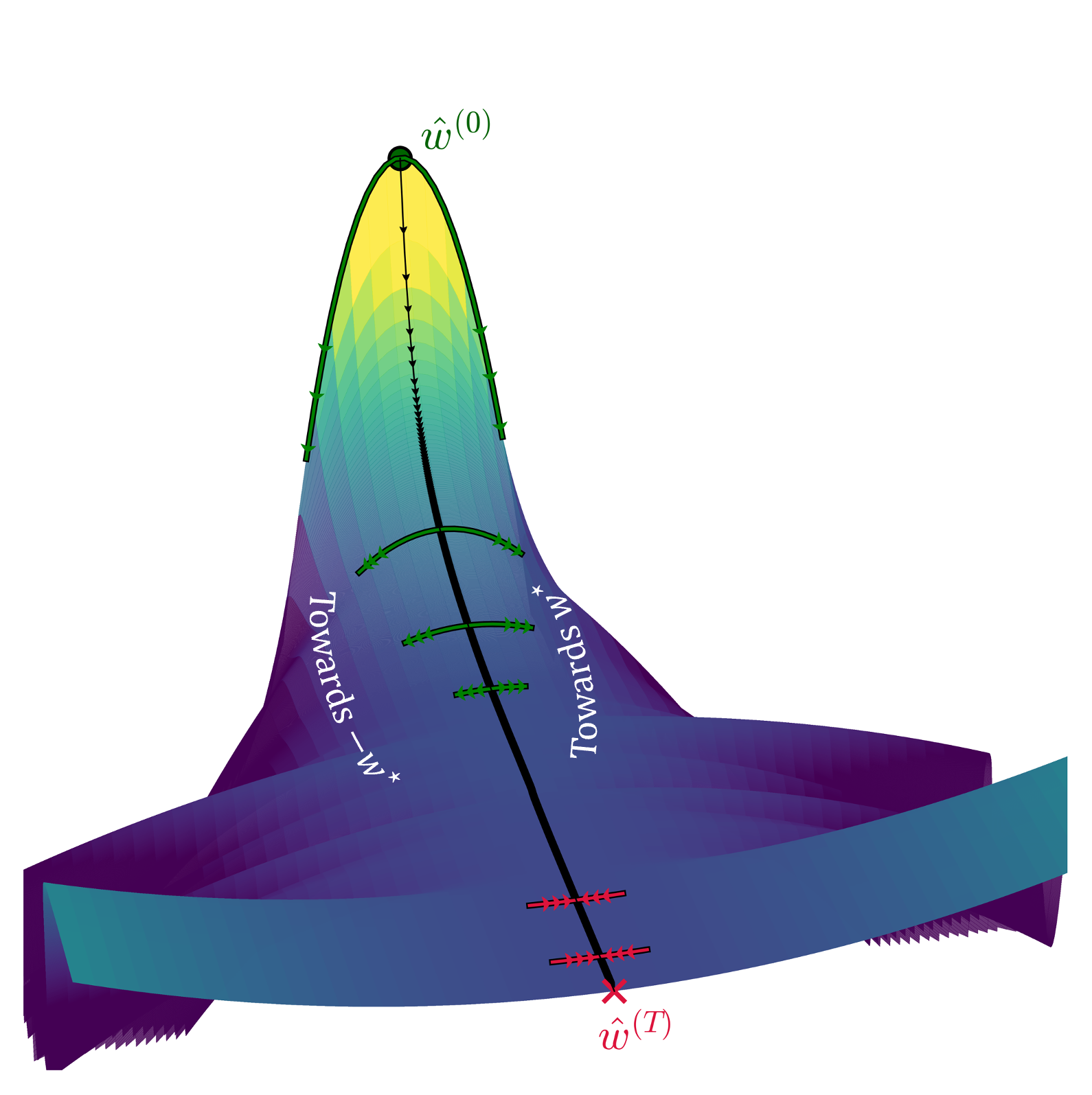}
    \caption{\textit{(Left)} Phases of the gradient flow dynamics in the phase retrieval loss landscape for $N\rightarrow \infty$. $\rho(\lambda)$ refers to the Hessian eigenvalue distribution, and the red bar shows when an outlier $\lambda_\star$ -- and hence a descent direction towards $\pm \bm{w}^\star$ -- exists. $t_\mathrm{TS}$ is the time required to reach a threshold state.
    \textit{(Right)} Evolution of the local curvature: dynamics projected in the direction of least stability $\bm{v}_1^{(t)}$ of the Hessian matrix (black arrows) in regime II for $N=256$. The green arrows indicate downward directions towards $\pm \bm{w}^\star$. At the end, the local curvature has become positive (red arrows).
    }
    \label{fig:summary_diagram}
\end{figure*}

\section{Summary of our contributions}

Although the spectral properties of the loss Hessian are conjectured to play a role during gradient descent, there are no analytical results characterizing the dynamical evolution of the Hessian and connecting it to the dynamics of the system. Here, we fill this gap focusing on phase retrieval as a non-convex problem in large dimensions $N,M \to \infty$ at fixed SNR $\alpha=M/N \sim O(1)$. We exhibit different regimes depending on $\alpha$ that are linked to the smallest eigenvalue $\lambda_1$ of the Hessian (summarized in the left panel of Figure~\ref{fig:summary_diagram}):
\begin{enumerate}[label=\Roman*.,topsep=2pt,itemsep=3pt,labelindent=20pt] 
    \item {\bf Rugged and featureless landscape:} When $\alpha < \alpha_\mathrm{BBP}^\mathrm{init}$, random initial conditions have no direction correlated with $\pm \bm{w}^\star$. The dynamics is unable to find back the signal and gets stuck into high-loss minima that are marginally stables (i.e., with a vanishing $\lambda_1$), the threshold states;
    \item {\bf Local curvature is informative \& dynamical transition in the Hessian:} When $\alpha \in \left[ \alpha_{\mathrm{BBP}}^\mathrm{init};\alpha_{\mathrm{BBP}}^{\mathrm{TS}}\right]$, the landscape at \textit{any} initial condition $\hat{\bm{w}}^{(0)}$ has a downward direction $\bm{v}_1^{(0)}$ aligned with $\pm \bm{w}^\star$.
    However, while descending, the direction $\bm{v}_1^{(t)}$ rotates away from the signal. At a finite time, a BBP transition takes place in the Hessian. In consequence, the correlation is lost, and the dynamics gets once again trapped into bad minima (threshold states);
    \item {\bf Full instability towards the signal:} When $\alpha > \alpha_{\mathrm{BBP}}^{\mathrm{TS}}$, the threshold states turn from local minima to saddle-points that have exactly one negative direction pointing towards the signal, making gradient descent escape the equator and converge to a well-generalizing (global) minimum in the second phase of the dynamics.
\end{enumerate}

\noindent These findings, that we obtain in the $N\to\infty$ limit, have crucial consequences for finite but large dimensions. 
In regime II, the local curvature towards the good minima is negative at the beginning of the dynamics and positive at the very end, as shown in the right panel of Figure~\ref{fig:summary_diagram}. Initially, the system has an overlap of order $1/\sqrt{N}$ with the signal and is able to escape the equator due to the initial negative curvature in a time of order $\log N$ \cite[see also][]{BenArous2021, bonnaire2023highdimensional, Arnaboldi2023}. Therefore, when $N\to\infty$ this timescale diverges, and the dynamics gets trapped in the threshold states at the equator before escaping and cannot recover the signal until $\alpha$ becomes larger than $\alpha_{\mathrm{BBP}}^{\mathrm{TS}}$ where they turn to saddles.
Nevertheless, for finite -- even very large -- $N$, the initial descent direction can be exploited to acquire, in the relatively short timescale $\log N$, a finite correlation with the signal. 
This finite-dimensional effect disappears only logarithmically with $N$ meaning it should lead to an effective transition growing with $\log N$.
Our analysis of the gradient-based dynamics at the equator of the phase retrieval landscape fills the gap between two earlier static studies. On one hand, \cite{Lu2020} characterizes the BBP transition for spectral methods, which can be seen as a particular case of the Hessian at initialization. On the other hand, \cite{Mannelli2020b} shows the existence of regimes I and III in the $N\to\infty$ limit. We therefore complete this picture by revealing a mechanism in the intermediate $\alpha$ regime that is driven by the local curvature of the Hessian and allows to amplify the small initial overlap to avoid bad minima in the landscape.
In particular, this phenomenon is very relevant for practical applications, and explains the large \emph{negative} gap reported in \cite{Mannelli2020b} between the SNR found numerically and theoretically.
Moreover this effect highlights why a good initialization -- in particular by spectral methods -- is critical for navigating rough and non-convex landscapes. We study phase retrieval precisely because it exhibits this intermediate regime, unlike simpler problems such as matrix-tensor PCA \cite{Mannelli2019a}.


\section{A motivating example} \label{sect:motiv}

\begin{figure}
    \centering
    \includegraphics[width=.45\linewidth]{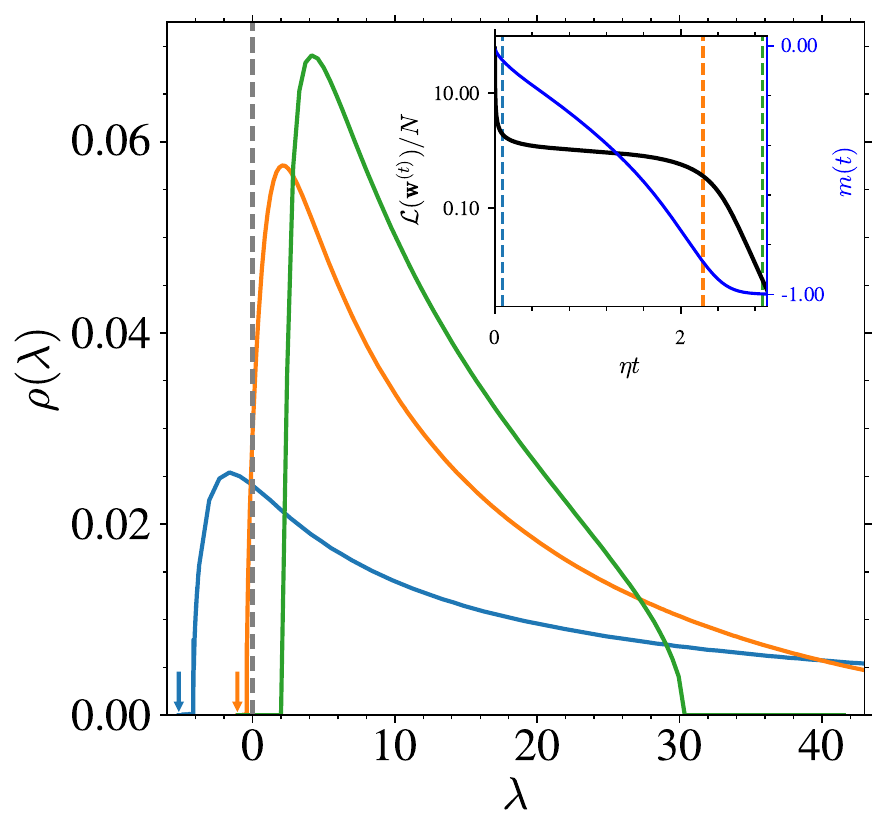}
    \caption{Hessian eigenvalue distribution $\rho(\lambda)$ of a simulation converging to $-\bm{w}^\star$ with $N = 2048$ and $\alpha = 3.1$. The inset shows the evolution of the rescaled loss function $\mathcal{L}(\hat{\bm{w}}^{(t)})/N$ (black curve) and the magnetization $m(t)$ (blue curve) with the simulation time $\eta t$. Colored lines in the main plot refer to different times shown in the inset. Arrows indicates the minimum eigenvalue when it is isolated from the bulk.}
    \label{fig:hessian_dynamics}
\end{figure}

To illustrate the phenomenon we will analyze later, let us examine a numerical example of a trajectory in the intermediate regime II. This example, displayed in Figure~\ref{fig:hessian_dynamics}, shows the evolution of the eigenspectrum at various timesteps during a successful gradient descent run initialized randomly with $\alpha = 3.1$ and $N=2048$. The inset highlights two dynamical regimes. First, the loss function quickly decreases to reach a plateau in which the system gets stuck for most of the simulation time. Second, a descent phase where the dynamics finally escapes the saddle-point and reaches zero loss. As the system gradually approaches a low-loss state, the Hessian displays a single negative eigenvalue in the direction of the signal (blue and orange arrows). As we will see analytically in Section~\ref{sect:RMT_BBP}, the local curvature towards the signal is negative from the very beginning of the dynamics (blue arrow). The system therefore exploits this direction before getting trapped in the threshold states that would be stable at this value of $\alpha$, and eventually reaches a global minimum with all positive eigenvalues (green curve), and a magnetization $m(T) = -1$. \\
The evolution of $m(t)$ in the inset -- growing while the loss is decreasing -- is due to the initial negative local curvature. It is exploited when $N$ is finite, allowing $m(t)$ to grow from its initial value $1/\sqrt{N}$ on timescales of order $\log N$. This mechanism moves the system away from the equator, in a region where the landscape is easier to descend \cite{Ros2019} which, in turn, enables gradient descent to succeed and find back the signal $-\bm{w}^\star$ despite being in a regime of $\alpha$ where bad minima are still present at the equator.

\section{Theory of the BBP transitions in the phase retrieval loss landscape} \label{sect:RMT_BBP}

\subsection{Hessian eigenspectrum and BBP condition}

We now present the theoretical framework allowing to study the spectral properties of the Hessian during the gradient descent dynamics. 
The Hessian matrix associated to the phase retrieval optimization is of the form
\begin{equation} \label{eq:hessian}
    \mathcal{H}(\hat{\bm{w}}^{(t)}) = \sum_{i=1}^M f(y_i, \hat{y}_i, t) \bm{x}_i \bm{x}_i\tran - \mu^{(t)} \bm{I}_N,
\end{equation}
with $f(y_i, \hat{y}_i, t) = \partial^2_{\hat{y}_i} \ell_a(y_i, \hat{y}_i)$, and $\bm{I}_N$ the identity matrix of size $N\times N$. Note that the dependence in $t$ of $f(y_i, \hat{y}_i, t)$ comes from $\hat{y}_i$ computed from $\hat{\bm{w}}^{(t)}$. In what follows, we omit the spherical constraint without any loss of generality since it simply induces a shift of the eigenvalues by $\mu^{(t)}$.
When considering the data vectors $\bm{x}_i$ as i.i.d. Gaussian, $\mathcal{H}(\hat{\bm{w}}^{(t)})$ is a random matrix drawn from what is called the \emph{non-white Wishart ensemble} \cite{Peche2006}.
We are interested in characterizing the value of $\alpha$ at which the smallest eigenvalue of the Hessian, detaches from the bulk and its associated eigenvector $\bm{v}_1^{(t)}$ has a finite scalar product with the signal. In this case, $\lambda_1^{(t)} = \lambda_\star^{(t)}$ creates an outlier as seen in of the left panel of Figure~\ref{fig:summary_diagram} (red bars). This transition of the smallest eigenvalue is called BBP transition \cite{Baik2005} and can be characterized analytically. Resorting to the tools from random matrix theory, we derive in  \ref{appendix:BBP_transition} equations for the behavior of the border of the bulk and the outlier eigenvalue $\lambda_\star^{(t)}$, when it exists at time $t$. The transition value of $\alpha$, referred to as $\alpha_{\mathrm{BBP}}$, satisfies
\begin{align} 
    \lambda_\star^{(t)} &= \alpha_\mathrm{BBP}(t) \mathbb{E}_{y, \hat{y}}\left[ \frac{f(y, \hat{y}, t) y^2}{1 - f(y, \hat{y}, t) \mathcal{S}_{-}(t)}\right], \label{eq:BBP_generic_1} \\
    \mathcal{S}_{-}(t) &= \alpha_\mathrm{BBP}(t) \mathbb{E}_{y, \hat{y}}\left[ \frac{f(y, \hat{y}, t)^2}{\left(1 - f(y, \hat{y}, t) \mathcal{S}_{-}(t)\right)^2}\right]. \label{eq:BBP_generic_2}
\end{align}
We provide in \ref{appendix:numerical_hessian} numerical evidence that these equations give accurate predictions of the spectrum of matrices in the form of \eqref{eq:hessian}, even at finite $N$.
When $\alpha > \alpha_\mathrm{BBP}(t)$, the eigenvector $\bm{v}_1^{(t)}$ associated to the smallest eigenvalue of the Hessian matrix $\mathcal{H}(\hat{\bm{w}}^{(t)})$ displays a non-zero overlap with the signal $\pm\bm{w}^\star$ that can be expressed (see \ref{appendix:overlap}) as
\begin{equation} \label{eq:overlap}
    (\bm{v}_1^{(t)} \cdot \bm{w}^\star)^2 = \frac{1}{1-\partial_z \Sigma(z)_{|z=\lambda_\star}},
\end{equation}
where
\begin{equation} \label{eq:Sigma_z}
    \Sigma(z) = \alpha \mathbb{E}_{y, \hat{y}}\left[\frac{f(y, \hat{y}, t) y^2}{1 - f(y, \hat{y}, t) \mathcal{S}_\mathcal{H}(z, t)}\right],    
\end{equation}
and
\begin{equation} \label{eq:Stieltjes_hessian}
    \mathcal{S}_\mathcal{H}(z, t)^{-1} = z - \alpha \mathbb{E}_{y, \hat{y}}\left[ \frac{f(y, \hat{y}, t)}{1 - f(y, \hat{y}, t) \mathcal{S}_\mathcal{H}(z, t)
}\right].
\end{equation}
The BBP condition, as well as the squared overlap $(\bm{v}_1^{(t)} \cdot \bm{w}^\star)^2$, are consequently expressed in terms of expectations computed over the joint probability distribution of the true and estimated labels at time $t$, namely $p(y, \hat{y}, t)$. Once it is known, one can solve the self-consistent equations~\eqref{eq:BBP_generic_1} and~\eqref{eq:BBP_generic_2} to obtain the value of $\alpha_\mathrm{BBP}(t)$, where $\bm{v}_1^{(t)}$ develops a non-zero correlation with the signal characterized by \eqref{eq:overlap}.
The rest of this section is devoted to analyze for which values of $\alpha$ and $t$ the BBP transition takes place at the equator of the phase retrieval loss landscape during the gradient descent dynamics.

\begin{figure}
    \centering
    \includegraphics[width=.50\linewidth]{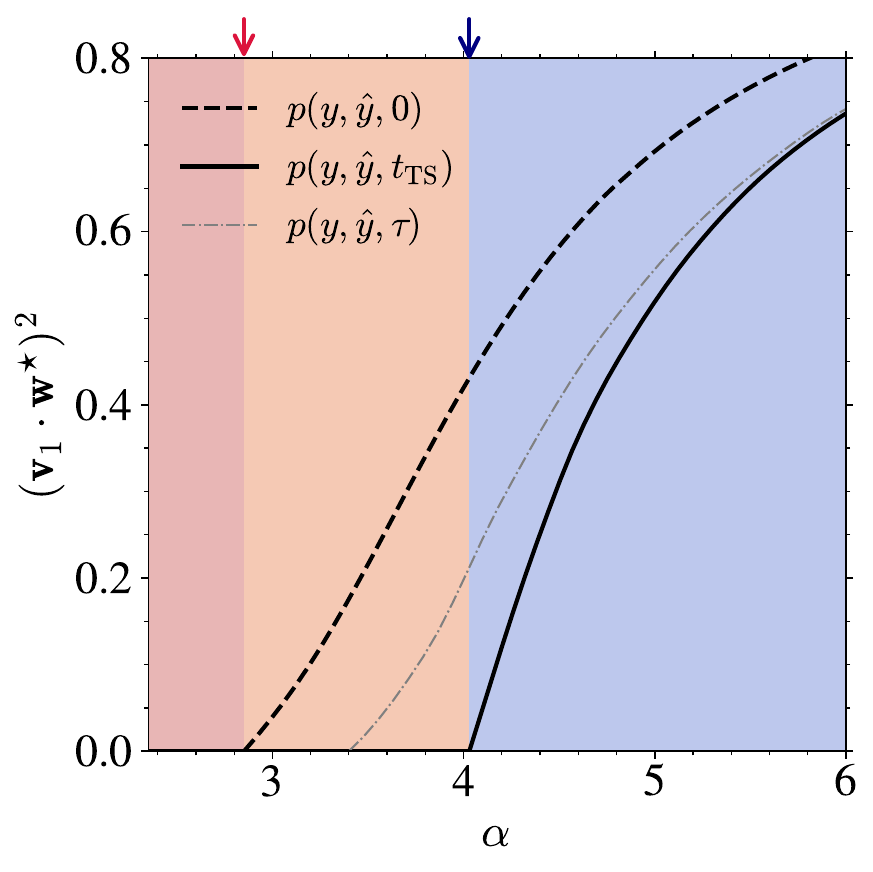}
    \includegraphics[width=.484\linewidth]{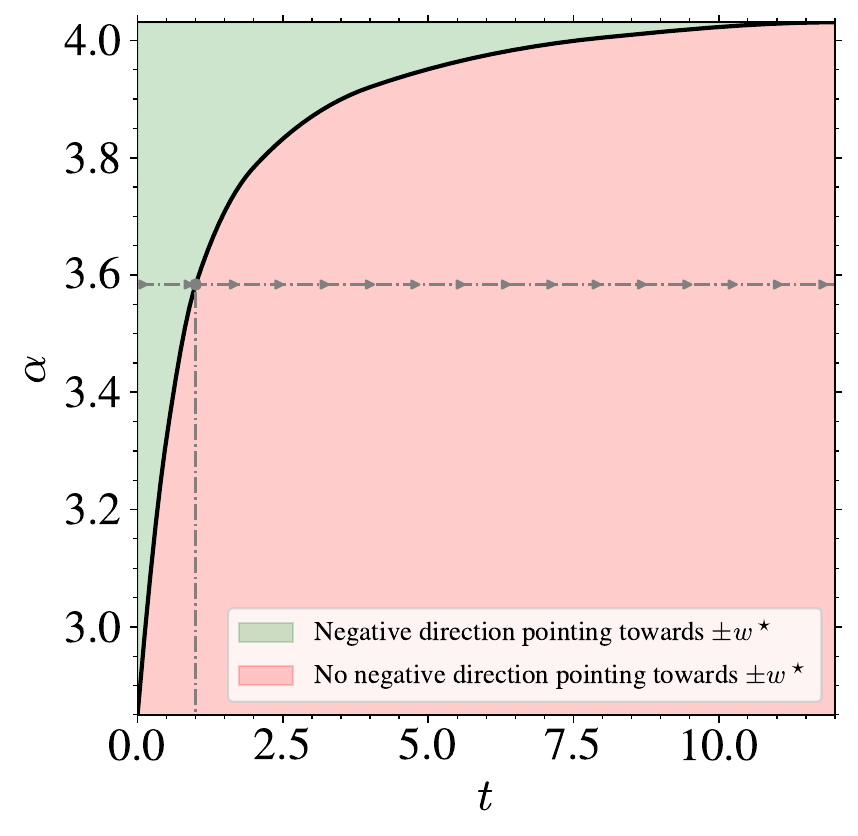}
    \caption{{Dynamical BBP theory at the equator of the phase retrieval loss landscape for $N\to\infty$.} \textit{(Left)} Evolution of $(\bm{v}_1^{(t)} \cdot \bm{w}^\star)^2$ from \ref{eq:overlap} at initialization ($t=0$, dashed line), on threshold states ($t=t_\mathrm{TS}$, solid line), and at an intermediary time $\tau \in \left[0, t_\mathrm{TS}=\infty\right]$ (grey dashed-dotted line). The red (resp. blue) arrow indicates $\alpha_\mathrm{BBP}^\mathrm{init} = 2.85$ (resp. $\alpha_\mathrm{BBP}^\mathrm{TS} = 4.03$). The background colors refer to the three different regimes introduced in Figure~\ref{fig:summary_diagram}. \textit{(Right)} Dynamical phase diagram of the value of $\alpha$ required for a BBP transition to take place with the descent time $t$. The grey dashed-dotted line refers to the intermediate time $\tau$ of the left panel. In both plots, the loss function is given by \eqref{eq:loss} with $a=0.01$.
    }
    \label{fig:overlaps}
\end{figure}

\subsection{BBP transition at initialization}
In the case of i.i.d. Gaussian measurements $\bm{x}_i$, and before operating gradient descent, $p(y, \hat{y}, t=0)$ is the product of two Gaussians. Solving the aforementioned equations characterizing the BBP transition grants the value $\alpha_\mathrm{BBP}^{(t=0)} \eqdef \alpha_\mathrm{BBP}^\mathrm{init} = 2.85$ for $\ell_{a=0.01}$. As a consequence, whenever $\alpha > \alpha_\mathrm{BBP}^\mathrm{init}$, any initial condition $\hat{\bm{w}}^{(0)}$ is characterized by a Hessian spectrum with an isolated left-most eigenvalue and an eigenvector $\bm{v}_1^{(0)}$ pointing towards $\pm \bm{w}^\star$. More precisely, $\bm{v}_1^{(0)}$ has a finite overlap with the signal that grows with $\alpha$, and which can be computed from \eqref{eq:overlap}. This evolution is displayed as the dashed line in the left panel of Figure~\ref{fig:overlaps}.

\subsection{BBP transition on threshold states}
The characterization of the joint probability on threshold states, denoted  $p(y, \hat{y}, t_\mathrm{TS})$, is more involved than at initialization. Note also that $t_\mathrm{TS} = \infty$ when $N\to\infty$ (and grows with $N$ when it is finite).
Right after a single step of gradient descent, $y$ and $\hat{y}$ are correlated. To pursue our analysis of the Hessian on these peculiar states, we employ two methods to approximate $p(y, \hat{y}, t_\mathrm{TS})$: (i) through adapted numerical simulations (described more precisely in Section~\ref{sect:Numerical}) sampling the threshold states. We then evaluate empirically the expectations in equations~\eqref{eq:BBP_generic_1} and~\eqref{eq:BBP_generic_2}; (ii) through the replica method from disordered systems (see \ref{appendix:replicas}), as performed in \cite{Franz2017, Mannelli2020b}.
Those two methods grant us two consistent but different values of the BBP transition on threshold states that are respectively $\alpha_\mathrm{BBP}^\mathrm{TS} = 4.03$ and $\alpha_\mathrm{BBP}^\mathrm{1RSB, TS} = 4.29$ for the loss \eqref{eq:loss} with $a=0.01$. We expect the gap between these two values to vanish when moving to higher order of replica symmetry breaking and we adopt $\alpha_\mathrm{BBP}^\mathrm{TS}$ as the BBP threshold for the rest of the paper.
For $\alpha > \alpha_\mathrm{BBP}^\mathrm{TS}$, also the threshold states turn from minima to saddles and develop a negative direction pointing towards $\pm \bm{w}^\star$: the overlap of the corresponding eigenvector with the signal is shown as the solid line in the left panel of Figure~\ref{fig:overlaps}.

\subsection{Dynamical BBP transitions}  \label{sect:finite_N_perspectives}
Comparing the evolution of the overlaps at $t=0$ and $t=t_\mathrm{TS}$ in the left panel of Figure~\ref{fig:overlaps}, we find that gradient descent transports the initial state towards a location that is in an even rougher part of the landscape, and that does \emph{not} allow recovery in the entire intermediate region of $\alpha \in \left[\alpha_\mathrm{BBP}^\mathrm{init},\, \alpha_\mathrm{BBP}^\mathrm{TS}\right]$, despite the initial local curvature at $t=0$. 
In this regime of SNR, and at a finite time $t_\mathrm{BBP}(\alpha)$, a BBP transition takes place during the descent as the informative isolated eigenvalue enters the bulk distribution, as illustrated by the horizontal dashed-dotted line in the right panel of Figure~\ref{fig:overlaps}.
The two ideal limits discussed above corresponds to $t_\mathrm{BBP}(\alpha_\mathrm{BBP}^\mathrm{init})=0$ and $t_\mathrm{BBP}(\alpha_\mathrm{BBP}^\mathrm{TS})=\infty$ but the same endeavor can be pursued for intermediate descent times using empirical expectations and numerical simulations.
For instance, Figure~\ref{fig:overlaps} reveals that for $\alpha \approx 3.57$, an initial negative local curvature pointing towards the signal exists and remains until a finite descent time of $t_\mathrm{BBP}(\alpha)=1$ after which it disappears, thus preventing signal recovery.

\subsection{Finite $N$ dynamics}
So far, the results of this section are obtained in the $N\to\infty$ limit. When $N$ is finite (but still large), the initial overlap is $m(t=0) \approx 1/\sqrt{N}$, as a consequence of the central limit theorem for large $N$. When $\alpha > \alpha_\mathrm{BBP}^\mathrm{init}$, the component along the signal direction grows exponentially due to the initial negative curvature, but with a prefactor $1/\sqrt{N}$, i.e. $m(t) \approx \exp\left(ct\right)/\sqrt{N}$ with $c$ a constant associated to $\lambda_1$. A time of order $\frac{1}{2c}\log N$ is hence needed to reach a magnetization of order one (and escape the equator in some cases), even for $\alpha<\alpha_\mathrm{BBP}^\mathrm{TS}$.
For $N\rightarrow \infty$ and $\alpha_\mathrm{BBP}^\mathrm{init}<\alpha<\alpha_\mathrm{BBP}^\mathrm{TS}$, this time diverges and the system looses the negative local curvature before actually being able to use it. This is for instance shown in the right panel of Figure~\ref{fig:overlaps} where the negative curvature towards $\pm\bm{w}^\star$ is lost in finite times. However, this happens only in the strict large $N$ limit. For finite (even very large) $N$, the situation changes substantially as $\log N$ is \emph{not} such a large timescale. In consequence, the system can acquire a magnetization of order one before hitting the dynamical BBP transition where the descent direction is lost, and hence avoid the bad minima of the equator. This should lead to a finite-$N$ algorithmic transition well below $\alpha_\mathrm{BBP}^\mathrm{TS}$, even for very large $N$.

This phenomenon plays a crucial role in practice by moving the system away from the equator during the descent, where the loss landscape is expected to become more benign \cite{Ros2019}, therefore enabling early-on successes when $N$ is finite. This was for instance hinted in Section~\ref{sect:motiv} and Figure~\ref{fig:hessian_dynamics}, where the inset shows the growth of the magnetization $m(t)$ from $1/\sqrt{N}$ at initialization to values of order one when the loss plateaus. As $\lvert m(t)\rvert$ increases, $\hat{\bm{w}}^{(t)}$ reaches a threshold state of large magnetization having a descending direction pointing to $-\bm{w}^\star$ (orange arrow) that is enabling recovery, despite the existence of bad minima at the equator that remain stables until $\alpha^\mathrm{TS}_\mathrm{BBP}$.
In the following, we test these hypotheses for finite $N$ through numerical experiments.


\section{Numerical analysis of the gradient descent dynamics} \label{sect:Numerical}

We run multiple experiments to analyze the behavior of gradient descent initialized both randomly and spectrally at finite $N$ by solving \eqref{eq:dynamics_GD} at fixed learning rate $\eta = 2\times10^{-4}$ for $T = 12,000\log_2(N)$ steps. We consider a system to perform strong recovery (meaning that $m(T)=\pm1$) whenever $\lvert m(T) \rvert \geq 0.99$. 

\begin{figure}
    \centering
    \begin{minipage}{0.453\textwidth}
        \centering
        \includegraphics[width=\linewidth]{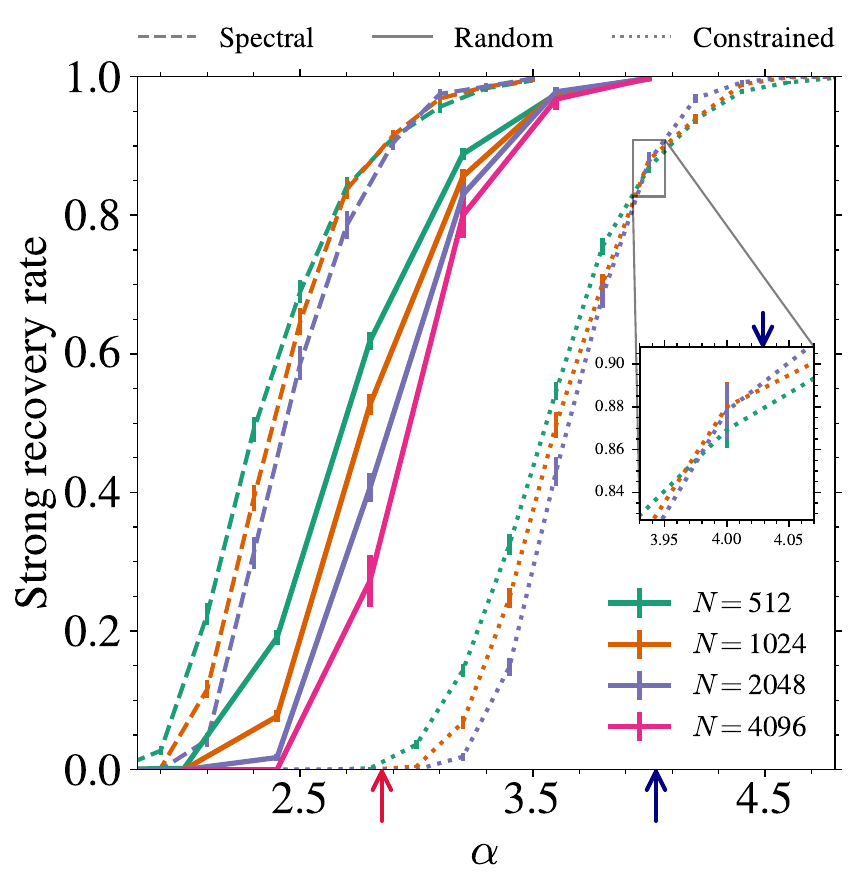}
        \caption{Strong recovery rates as a function of $\alpha$ for different $N$ and initialization schemes: spectral, random, and constrained.
        The red (resp. blue) arrow indicates $\alpha_\mathrm{BBP}^\mathrm{init}$ (resp. $\alpha_\mathrm{BBP}^\mathrm{TS}$). The error bars represent $95\%$ confidence intervals on the mean.
        }
        \label{fig:fraction_success}
    \end{minipage}\hfill
    \begin{minipage}{0.473\textwidth}
    \centering
    \includegraphics[width=\linewidth]{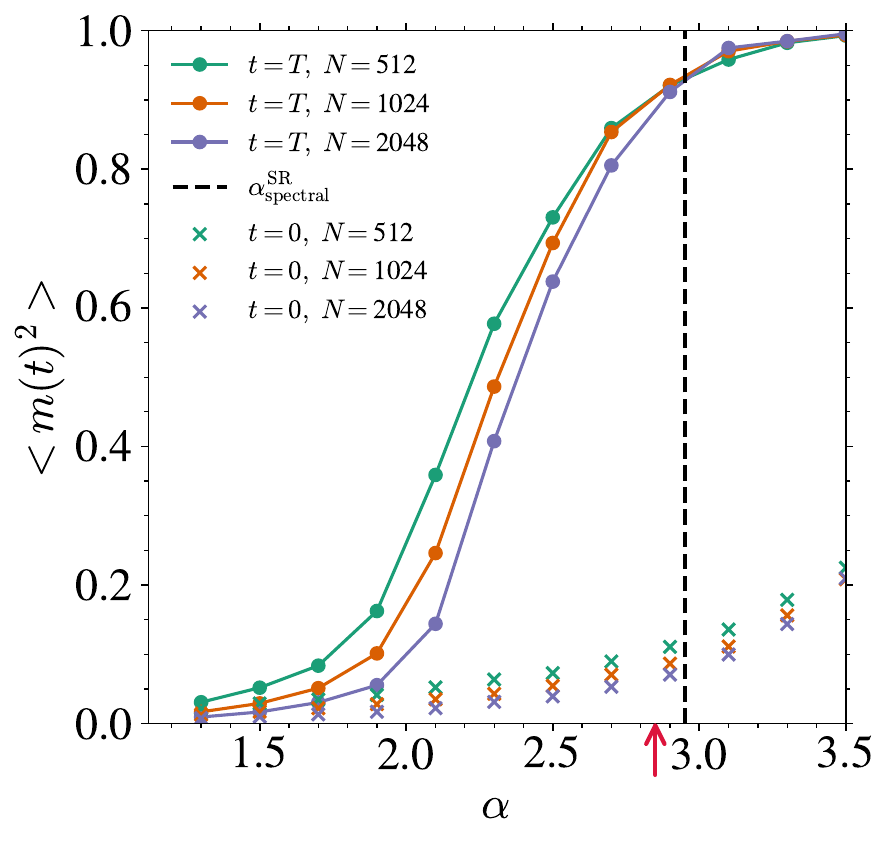}
    \caption{Evolution of the averaged squared magnetization $\langle m(t)^2 \rangle$ with $\alpha$ for several values of $N$ at times $0$ and $T$ using spectral initialization along $\bm{v}_1^{(0)}$. The red arrow denotes $\alpha_\mathrm{BBP}^\mathrm{init}$ and the vertical dashed black line corresponds to the strong recovery threshold $\alpha_\mathrm{spectral}^\mathrm{SR}$.}
    \label{fig:Spectral_mag}
    \end{minipage}
\end{figure}

First, let us focus on randomly initialized weights $\hat{\bm{w}}^{(0)} \sim \mathcal{N}(\bm{0}_N, \bm{I}_N)$, leading to the strong recovery rates shown as solid lines in Figure~\ref{fig:fraction_success} for $N \in \{512, 1024, 2048, 4096\}$. In agreement with the previous arguments, the simulations achieve strong recovery well before $\alpha_\mathrm{BBP}^\mathrm{TS}$, without clearly intersecting each other. This gap between the simulations and theoretical BBP prediction was also observed in \cite{Mannelli2020b}. As discussed in Section~\ref{sect:finite_N_perspectives}, it is due to the displacement of the effective transition logarithmically with $N$ (see also \ref{appendix:numerical}).
In what follows, we devise more elaborated ways of exploring the landscape to avoid that this finite size effect draws the dynamics away from the bad minima before the actual BBP transition.

\subsection{A constrained optimization to probe threshold states} \label{subsect:constrained}

Efficiently sampling the threshold states numerically at finite $N$ is a critical aspect of our numerical analysis to show that:
\begin{enumerate}[label*=\arabic*.,topsep=2pt,itemsep=3pt,labelindent=20pt]
    \item These states exist in the phase retrieval loss landscape,
    \item Gradient descent is trapped into them when $\alpha < \alpha_\mathrm{BBP}^\mathrm{TS}$,
    \item They are responsible for the BBP transition at the end of the dynamics for $N\to\infty$.
\end{enumerate}
In order to sample the threshold states, we constrain the optimization to remain at the equator by projecting the estimate at each time step $t$ in the subspace orthogonal to $\bm{w}^{\star}$,
\begin{equation}
    \hat{\bm{w}}^{(t)}_\perp = \left(\bm{I}_N - \frac{{\bm{w}^\star} {\bm{w}^\star}\tran}{N}\right) \hat{\bm{w}}^{(t)},
\end{equation}
where $\hat{\bm{w}}^{(t)}$ is defined in \eqref{eq:dynamics_GD}. While sticking to the equator, the loss is still gradually decreased until it reaches a plateau as in Figure~\ref{fig:hessian_dynamics}, but with an enforced magnetization of zero. In practice, we perform $t_\mathrm{c} = 60,000$ gradient descent steps with the constraint and converge to a state $\hat{\bm{w}}^{(t_\mathrm{c})}$ that we use as initialization for the standard (unconstrained) gradient descent, resulting in a procedure called \emph{constrained initialization}. More details about this procedure and the algorithm can be found in~\ref{appendix:numerical}.
Although this numerical scheme is not properly speaking sampling the threshold states since the gradient cannot be zero in the direction of the signal, its component is $\sqrt{N}$ smaller than the gradient norm. We have checked numerically that the states we visit have the expected properties (marginal Hessian, BBP transition, and eigenvalues distribution).

We show as the dotted lines in Figure~\ref{fig:fraction_success} the strong recovery rates obtained with constrained initialization. Contrary to what was observed in the case of random initialization, the successes for different values of $N$ now seem to converge at around $\alpha_\mathrm{cons.}^\mathrm{SR} \approx 4.0$, in agreement with our theory from Section~\ref{sect:RMT_BBP}, and considerably shifting the success rates to larger $\alpha$ with respect to the random initialization case.
This means in particular that the threshold states exist at the equator and they indeed are found in a rougher part of the landscape, making it harder to converge to a well-generalizing minimum. By increasing the value of $a$ in the loss~\eqref{eq:loss}, we observe significant discrepancies between the predicted values for the BBP on threshold states and $\alpha_\mathrm{cons.}^\mathrm{SR}$ obtained numerically (see \ref{appendix:loss}). We leave to future works to solve this gap which could be due to additional -- and unidentified -- strong finite size effects.

\subsection{Spectral initialization, weak recovery and loss landscape away from the equator} \label{subsect:spectral}

As stated in Section~\ref{sect:RMT_BBP}, when $\alpha > \alpha_\mathrm{BBP}^\mathrm{init}$, the Hessian matrix of any random configuration $\hat{\bm{w}}^{(0)}$ has a direction of least stability $\bm{v}_1^{(0)}$ displaying a non-zero overlap with the signal. This idea is at the heart of what is called \textit{spectral initialization} proposed and studied in many previous works \cite{Candes2015, Wang2017a, Barbier2019, Mondelli2019, Lu2020, Luo2020, maillard2020phase, maillard2022construction}. By initializing the descent at $\hat{\bm{w}}^{(0)} = \bm{v}_1^{(0)}$, one expects the system to avoid the bad minima, or at least to reach threshold states of larger latitudes that may exhibit a BBP transition at a lower signal-to-noise ratio $\alpha$. From the perspective discussed in the previous sections, initializing along $\bm{v}_1^{(0)}$ is like taking advantage of the negative local curvature from the beginning of the dynamics.
The dashed lines of Figure~\ref{fig:fraction_success} support numerically these intuitions with a transition now occurring around $\alpha^\mathrm{SR}_\mathrm{spectral} \approx 2.95 < \alpha^{\mathrm{SR}}_\mathrm{random}$.
This is also emphasized by Figure~\ref{fig:Spectral_mag} in which we plot $\langle m(t)^2\rangle$ both at initialization along $\bm{v}_1^{(0)}$ (crosses) and after $T$ steps of gradient descent (dots and solid lines).

There are several important findings associated to Figure~\ref{fig:Spectral_mag}. First, there is a regime $\alpha<\alpha^\mathrm{SR}_\mathrm{spectral}$ in which the Hessian initialization leads to weak recovery (meaning it reaches states that have a finite magnetization $\lvert m(T) \rvert < 0.99$ in practice), and a regime $\alpha>\alpha^\mathrm{SR}_\mathrm{spectral}$ in which it leads to strong recovery ($\lvert m(T) \rvert \geq 0.99$). This phenomenon is actually more prominent for larger values of $a$ (see \ref{appendix:loss}). It hints at a complex characterization of the loss landscape away from the equator, with minima trapping the dynamics at low $\alpha$ but having a finite magnetization, see \cite{Mignacco2021} for related results and \cite{Ros2019} for a Kac-Rice perspective on simpler models. Second, Figure~\ref{fig:Spectral_mag} shows that by using the initial local negative curvature the system can achieve strong recovery well below $\alpha_\mathrm{BBP}^\mathrm{TS}$. These results therefore highlight the importance of a good initialization for gradient descent dynamics, especially when the landscape is more benign at the beginning of the dynamics than later on. 

\section{Discussion and perspectives}

We provide a theoretical study of the behavior of gradient flow in a high-dimensional and non-convex landscape through the Gaussian noiseless phase retrieval problem in a teacher-student setup. Based on the analytical and dynamical description of the Hessian spectrum during the dynamics, we are able to understand the main conditions of success and failure as a function of the signal-to-noise ratio $\alpha$. From this analysis, we draw several conclusions and perspectives at both finite and infinite $N$.

\paragraph*{\bf The local landscape is more benign and informative at the beginning of the dynamics.} The value of $\alpha$ required to induce a BBP transition in the Hessian matrix is larger on threshold states than at random initialization. However, for $N\rightarrow\infty$, although there exists one descending direction going towards $\pm \bm{w}^\star$ at $t=0$, gradient descent ignores it and ends up being trapped in the threshold states when $\alpha \in \left[\alpha_\mathrm{BBP}^\mathrm{init}, \alpha_\mathrm{BBP}^\mathrm{TS}\right]$. A larger signal-to-noise ratio $\alpha >\alpha_\mathrm{BBP}^\mathrm{TS}$ is then required to render the latter unstable.

\paragraph*{\bf Finite $N$ random initializations benefit from this phenomenon.} Due to the initial local curvature towards $\pm \bm{w}^\star$ existing at $\alpha > \alpha_{\mathrm{BBP}}^\mathrm{init}$, and to the finite value of $N$ used in practice, the magnetization $m(t)$ between the estimate and the signal is able to grow during the descent. This enables the system to escape the equator on a timescale of order $\log N$ by leaving the roughest part of the landscape and join more benign regions. This is \emph{the} mechanism that allows for successful optimization in practice, well before the algorithmic threshold corresponding to the high-dimensional limit $N \to \infty$.

\paragraph*{\bf The importance of spectral initializations.}
Given that the landscape is more benign at the beginning of the dynamics, spectral initializations can be very useful to escape the equator more efficiently before reaching bad and rougher regions. This phenomenon provides a showcase for a strong advantage of spectral initializations and, more generally, of spectral properties to improve optimization in non-convex and high-dimensional landscape -- a research direction that received a lot of attention recently in the context of deep learning \cite{Ghorbani2019, Sun2020optimization, Adahessian2021}. Our theoretical analysis of the BBP transitions holds at the equator, where $m=0$. To get a better understanding of spectral initializations, one must study the topological properties of the landscape as a function of both $\alpha$ and $m$. This could be done using the Kac-Rice method for loss functions in the form of \eqref{eq:loss} as proposed in \cite{Maillard2019b}.

\paragraph*{\bf Not all loss functions are equal.} The values of $\alpha$ at which the dynamical BBP transitions occur depend strongly on the choice of the loss function. Thus, it would be interesting to find losses that enhance this phenomenon and lead to an earlier signal recovery, as done in \cite{Mondelli2019} for spectral initializations and in \cite{Cai2022, Cai2023} for landscape trivialization. Finally, it would be worth characterizing this phenomenon for a broader class of loss functions. We show a first case study by varying $a$ in \eqref{eq:loss} in \ref{appendix:loss}.

\ack{The authors thank Stefano Sarao Mannelli for sharing his code used in \cite{Mannelli2020b}. T.B. further thanks Aurélien Decelle and Bruno Loureiro for useful discussions on the topic. G.B. acknowledges support from the French government under the management of the Agence Nationale de la Recherche as part of the ``Investissements d’avenir'' program, reference ANR-19-P3IA0001 (PRAIRIE 3IA Institute).
C.C. acknowledges financial support from PNRR MUR project PE0000013-FAIR and from MUR through PRIN2022 project 202234LKBW-Land(e)scapes. \\
The manuscript \cite{BenArous2025_local}, which appeared after completion and submission of our work, identifies a similar dynamical BBP mechanism for a broad family of losses during SGD dynamics. It also highlights the importance of the BBP transition during the training dynamics (online SGD for \cite{BenArous2025_local}, gradient flow in our case).}

\appendix
\setcounter{section}{0}

\section{Random matrix analysis of the Hessian} \label{appendix:hessian}

\subsection{Characterization of the Hessian spectrum and BBP transition} \label{appendix:BBP_transition}

Omitting the spherical constraint, which is just a translation of the eigensupport, and dropping the dependence in $t$ to lighten the notations, the Hessian matrix can be written as
\begin{equation}
    \mathcal{H}(\bm{w}^{(t)}) = \sum_{i=1}^M f(y_i, \hat{y}_i) \bm{x}_i \bm{x}_i\tran.
\end{equation}
We first focus on describing the bulk by neglecting the signal part. We want to obtain a self-consistent equation in the large $N$ limit on the Stieltjes transform of $\mathcal{H}$, denoted $S_\mathcal{H}(z) = \Tr \bm{G}/ N$, with $\bm{G}=(z\bm{I} - \mathcal{H})^{-1}$ the resolvent matrix. For this, we rely on the following simple identity
\begin{equation} \label{eq:simple_identity}
    (z\bm{I} - \mathcal{H}) \bm{G} = \bm{I},
\end{equation}
leading, after rearranging, taking the trace, and dividing by $N$, to
\begin{equation} \label{eq:master}
    z S_\mathcal{H}(z) = 1 +\frac{1}{N} \sum_{i=1} f(y_i, \hat{y}_i) \bm{x}_i\tran \bm{G} \bm{x}_i
\end{equation}
Since $\bm{G}$ depends on $\bm{x}_i$ through $\mathcal{H}$, we cannot simply reduce the quadratic form $\bm{x}_i\tran \bm{G}\bm{x}_i$ to the trace of $\bm{G}$. To compute this quantity, we therefore employ the \emph{cavity method} and consider a system made of $N-1$ particles satisfying
\begin{equation} \label{eq:cavity}
    \mathcal{H}_{-i} = \mathcal{H} - f(y_i, \hat{y}_i) \bm{x}_i \bm{x}_i\tran,
\end{equation}
meaning we remove the contribution of the $i^\mathrm{th}$ matrix in the Hessian. We can then link the resolvent matrices from the two systems using the Sherman-Morrison identity as
\begin{equation} \label{eq:Sherman-Morrison}
    \bm{G} = \bm{G}_{-i} + f(y_i, \hat{y}_i) \frac{\bm{G}_{-i} \bm{x}_i\bm{x}_i\tran \bm{G}_{-i}}{1 - f(y_i, \hat{y}_i) \bm{x}_i\tran \bm{G}_{-i} \bm{x}_i},
\end{equation}
meaning 
\begin{equation}
    \bm{x}_i\tran \bm{G} \bm{x}_i = \bm{x}_i\tran \bm{G}_{-i} \bm{x}_i + f(y_i, \hat{y}_i) \frac{\left(\bm{x}_i\tran \bm{G}_{-i} \bm{x}_i\right)^2}{1 - f(y_i,\hat{y}_i)\bm{x}_i\tran \bm{G}_{-i}\bm{x}_i}.
\end{equation}
Since $\bm{G}_{-i}$ is independent of $\bm{x}_i$ by construction, we can now use concentration arguments to write that $\bm{x}_i\tran \bm{G}_{-i}\bm{x}_i \approx \Tr \bm{G}_{-i} \mathbb{E}\left(\bm{x}_i\tran\bm{x}_i\right) = \Tr \bm{G}_{-i}/N \approx S_{\mathcal{H}_{-i}}(z)$. Finally, we use the cavity assumption that, in the the large $N$ limit, $S_{\mathcal{H}_{-i}}(z) \approx S_\mathcal{H}(z)$, to obtain
\begin{align}
    \bm{x}_i\tran \bm{G} \bm{x}_i &\approx S_\mathcal{H}(z) + f(y_i, \hat{y}_i) \frac{S_\mathcal{H}(z)^2}{1-f(y_i, \hat{y}_i) S_\mathcal{H}(z)}, \\
    &\approx \frac{S_\mathcal{H}(z)}{1-f(y_i, \hat{y}_i) S_\mathcal{H}(z)}.
\end{align}
Injecting it back into~\ref{eq:master}, we obtain the following self-consistent equation on the Stieltjes transform of the bulk part:
\begin{align}
    S_\mathcal{H}^{-1} &= z - \frac{1}{N}\sum_{i=1}^M \frac{f(y_i, \hat{y}_i)}{1-f(y_i, \hat{y}_i)S_\mathcal{H}}, \\
    &= z - \alpha \mathbb{E}_{y, \hat{y}}\left[ \frac{f(y, \hat{y})}{1-f(y, \hat{y})S_\mathcal{H}} \right],
\end{align}
where the expectation is taken over the joint probability distribution at time $t$ of $y$ and $\hat{y}$ that we denote $p(y, \hat{y}, t)$.
This equation fully characterizes the bulk of the eigenspectrum through the Sokhotski–Plemelj inversion formula allowing to recover the density of eigenvalues $\rho(\lambda)$, as used to obtain Figure~\ref{fig:hessian_dynamics}.

As argued in the main text, in the presence of an outlier eigenvalue due to the signal, the Hessian can be written as a sum of two contributions: one component independent from the signal -- the continuous bulk characterized by \eqref{eq:Stieltjes_hessian} -- and another component aligned with the signal. In particular, we can decompose the feature vectors as
\begin{equation}
    \bm{x}_i = \frac{y_i\bm{w}^\star}{N} + \bm{u}_i,
\end{equation}
where $\bm{u}_i \perp \bm{w}^\star$. 
To obtain a BBP condition for the Hessian of the phase retrieval loss landscape, we look for an eigenvalue creating a singularity in the full resolvent matrix in the signal direction ${\bm{w}^\star}\tran \bm{G} \bm{w}^\star$. For simplicity, and since the problem is invariant by rotation, let us assume without loss of generality that $\bm{w}^\star = \sqrt{N}\bm{e}_1$, where $\bm{e}_1 = \left[1, 0, \cdots, 0\right]\tran$.
From there, using~\eqref{eq:simple_identity}, we obtain the following expression for $g_{11} = \bm{e}_1\tran \bm{G} \bm{e}_1$,
\begin{equation} \label{eq:master_signal}
    z g_{11} = 1 + \sum_{i=1}^M f(y_i, \hat{y}_i) \bm{x}_i\tran \bm{e}_1 \left(\bm{x}_i\tran \bm{G}\bm{e}_1\right).
\end{equation}
It now remains two terms to evaluate. First, remark that
\begin{align}
    \bm{x}_i\tran \bm{e}_1 &= \left(\frac{y_i \bm{w}^\star}{N} + \bm{u}_i\right)\tran \bm{e}_1, \\
    &= \frac{y_i\sqrt{N}\bm{e}_1\tran \bm{e}_1}{N}, \\
    &= \frac{y_i}{\sqrt{N}}.
\end{align}
For the second term, we can use the cavity framework from~\eqref{eq:cavity} followed by the perturbative expansion of the Sherman-Morrison~\eqref{eq:Sherman-Morrison} to write
\begin{equation}
    \bm{x}_i\tran \bm{G}\bm{e}_1 = \frac{\bm{x}_i\tran \bm{G}_{-i} \bm{e}_1}{1 - f(y_i, \hat{y}_i) \bm{x}_i\tran \bm{G}_{-i} \bm{x}_i}.
\end{equation}
The quadratic form of the denominator $\bm{x}_i\tran \bm{G}_{-i} \bm{x}_i$ is dominated by the contributions of the perpendicular terms $\bm{u}_i$, meaning it can be approximated by the previous derivation for the bulk in the large $N$ limit, i.e. by $\Tr \bm{G}_{-i} / N \approx S_\mathcal{H}(z)$. For the numerator, we have
\begin{align}
    \bm{x}_i\tran \bm{G}_{-i} \bm{e}_1 &= y_i \frac{w^\star}{N} \bm{G}_{-i} \bm{e}_1 + \bm{u}_i \bm{G}_{-i}\bm{e}_1, \\
    &= \frac{y_i}{\sqrt{N}} \left[\bm{G}_{-i}\right]_{11}, \\
    &\approx \frac{y_i}{\sqrt{N}} g_{11},
\end{align}
where the first equality uses the definition of $\bm{x}_i$, the second the definition of $\bm{w}^\star$ and its orthogonality to $\bm{u}_i$, and the third the cavity approximation. This finally gives
\begin{equation}
    \bm{x}_i\tran \bm{G} \bm{e}_1 = \frac{y_i g_{11}}{\sqrt{N} \left(1 - f(y_i, \hat{y}_i)\right) S_\mathcal{H}(z)},
\end{equation}
that we can inject into~\eqref{eq:master_signal} to get 
\begin{equation}
    g_{11}^{-1} = z - \alpha \mathbb{E}_{y, \hat{y}}\left[ \frac{f(y, \hat{y}) y^2}{1-f(y,\hat{y})S_\mathcal{H}(z)} \right].
\end{equation}
Therefore, an outlier exists for $z = \lambda_\star$ satisfying
\begin{equation} \label{eq:Stieltjes_hessian_signal}
    \lambda_\star = \Sigma(\lambda_\star),
\end{equation}
with $\Sigma(z)$ defined in \eqref{eq:Sigma_z} as 
\begin{equation}
    \Sigma(z) = \alpha \mathbb{E}_{y, \hat{y}}\left[\frac{f(y, \hat{y}) y^2}{1 - f(y, \hat{y}) \mathcal{S}_\mathcal{H}(z)}\right].    
\end{equation}
This holds as long as $\lambda_\star < \lambda_{-}$, with $\lambda_{-}$ the left edge of the continuous part of the spectrum. A condition on $\lambda_{-}$ can be found through the maximum of $z(\mathcal{S})$, satisfying
\begin{equation}
    \frac{\partial z(\mathcal{S})}{\partial \mathcal{S}}_{\rvert_{\mathcal{S} = \mathcal{S}_{-}}} = 0.
\end{equation}
Since $z(\mathcal{S}) = R_\mathcal{H}(\mathcal{S}) + 1/\mathcal{S}$, we find that
\begin{equation} \label{eq:left_edge}
    \mathcal{S}_{-} = \left( \mathbb{E}_{y, \hat{y}}\left[\frac{\alpha^2 f(y, \hat{y})^2}{\left(1 - f(y, \hat{y}) \mathcal{S}_{-}\right)^2}\right]\right)^{-1/2},
\end{equation}
which is the condition of the Stieltjes transform of the left edge. Finally, by equating the left edge and the outlier eigenvalue equations, we obtain the BBP condition from \eqref{eq:BBP_generic_1}, i.e.
\begin{equation}
    \lambda_\star = \alpha_\mathrm{BBP} \mathbb{E}_{y, \hat{y}}\left[ \frac{f(y, \hat{y}) y^2}{1 - f(y, \hat{y}) \mathcal{S}_{-}}\right].
\end{equation}

\begin{figure*}
    \begin{center}
    \centerline{\includegraphics[width=\linewidth]{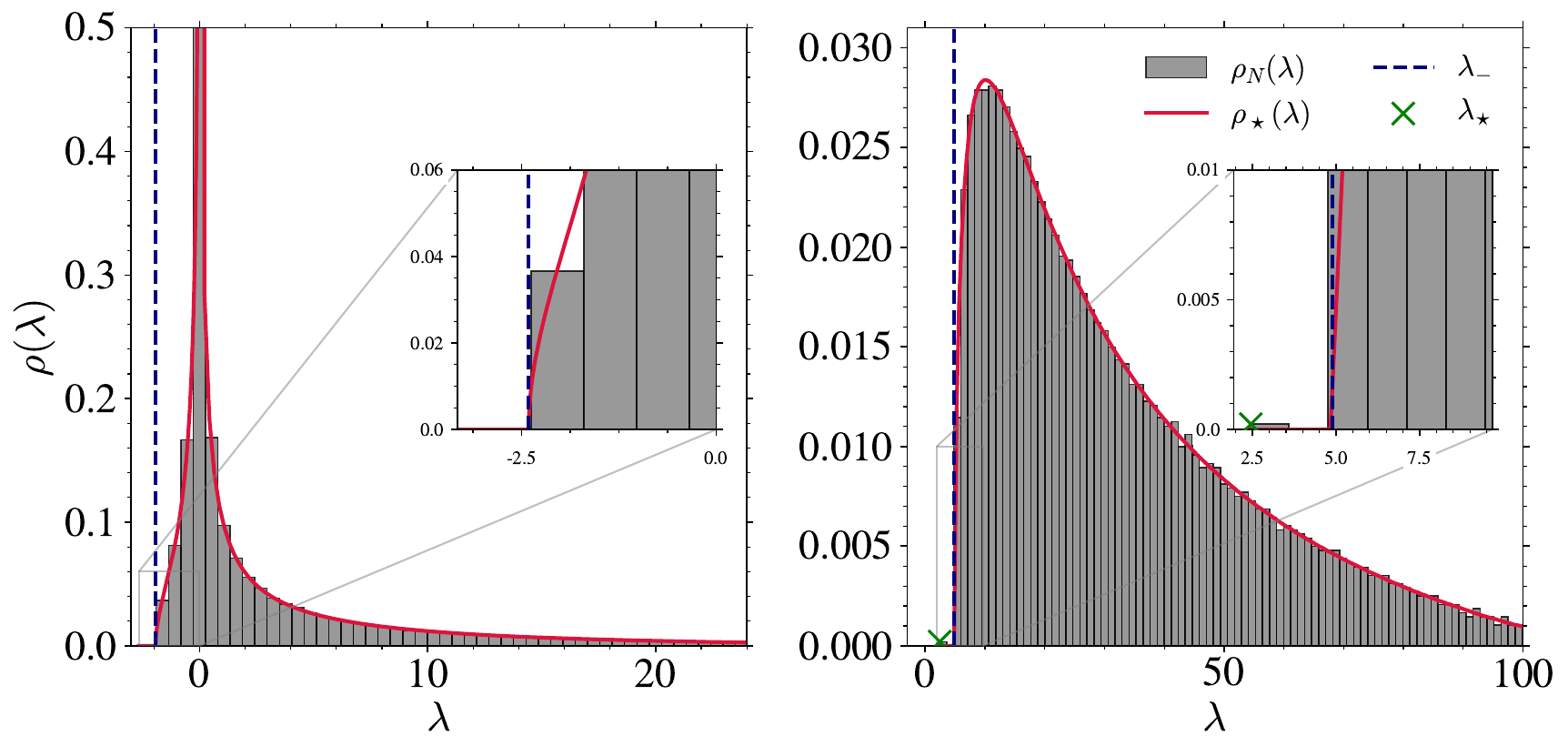}}
    \vspace{-.4cm}
    \caption{Illustrative comparison of the eigenspectrum properties analytically predicted from equations~\eqref{eq:Stieltjes_hessian},~\eqref{eq:BBP_generic_1}, and~\eqref{eq:BBP_generic_2} with empirical spectra. Eigenvalues are obtained at $t=0$ (initialization) for $N=4096$, $a=1$, and \textit{(Left)} $\alpha = 1$ or \textit{(Right)} $\alpha = 10$. The blue vertical dashed line shows the left edge estimation while the green cross indicates the outlier eigenvalue when it exists.}
    \label{fig:hessian_spectrum}
    \end{center}
    \vskip -0.2in
\end{figure*}

\begin{figure*}
    \begin{center}
    \centerline{
    \includegraphics[width=.49\linewidth]{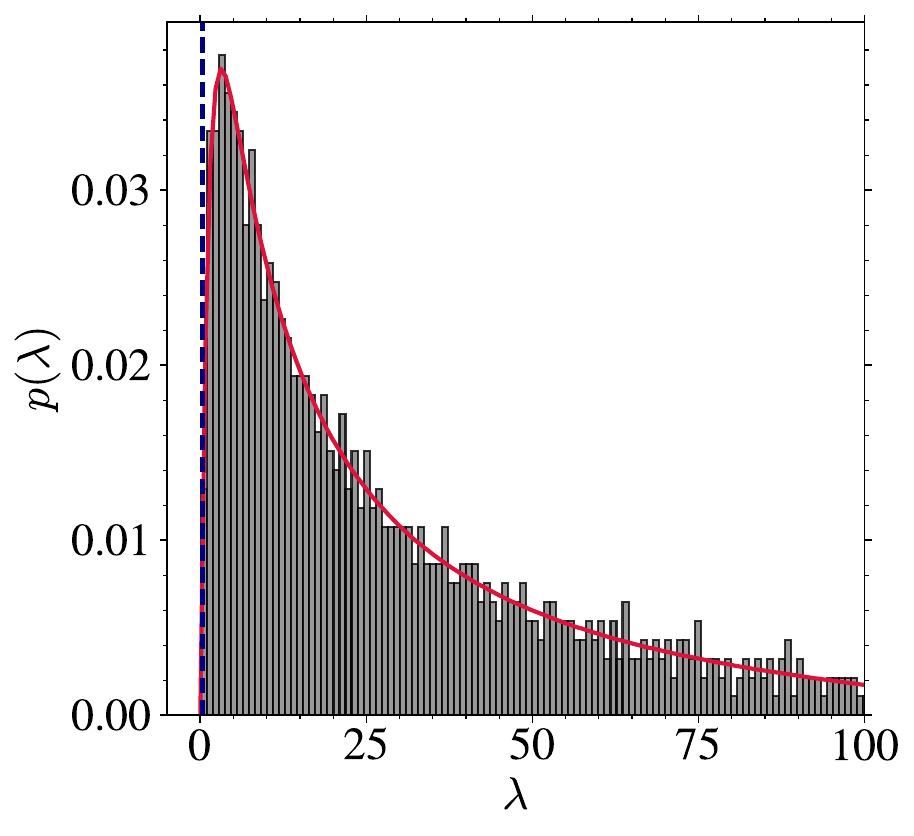}
    \includegraphics[width=.49\linewidth]{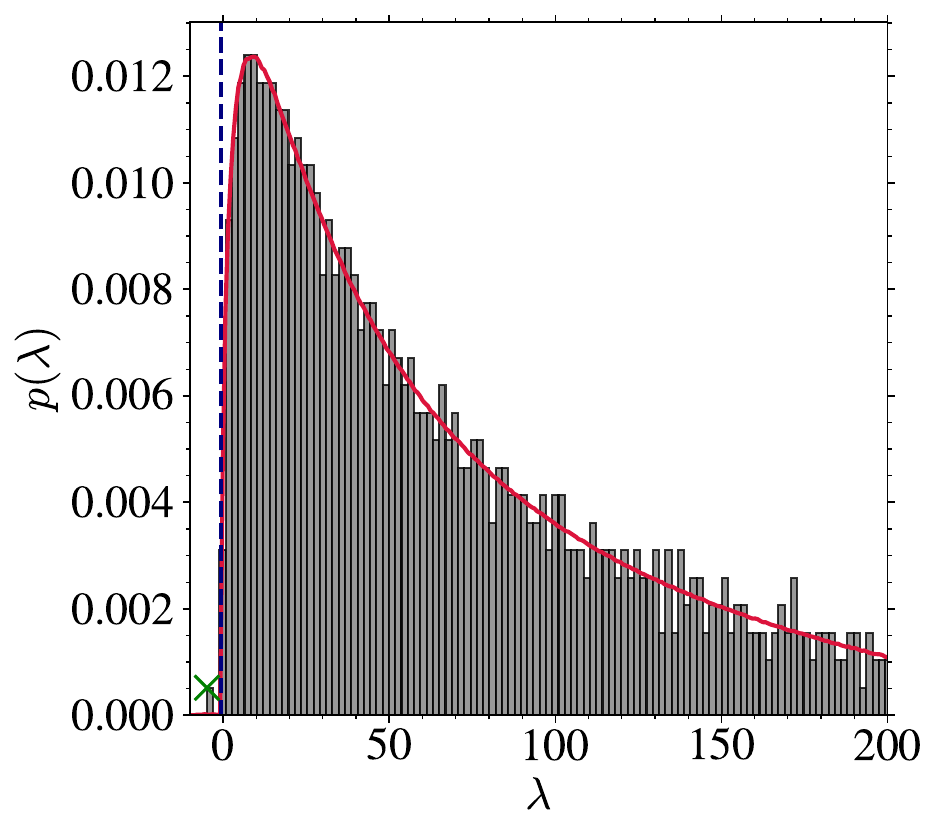}}
    \vspace{-.4cm}
    \caption{Same as Fig.~\ref{fig:hessian_spectrum} for intermediate times during the constrained initialization of gradient descent with $N=1024$, $a=0.01$, and \textit{(Left)} $\alpha = 3.6, \; t=4$; and \textit{(Right)} $\alpha = 7, \; t=8$.}
    \label{fig:hessian_spectrum_interm_times}
    \end{center}
    \vskip -0.2in
\end{figure*}

\subsection{Numerical validation of the random matrix equations} \label{appendix:numerical_hessian}

Let us now check the accuracy of our approach through a numerical experiment. Figure~\ref{fig:hessian_spectrum} shows two realizations of matrices in the form of \eqref{eq:hessian} for $N = 4096$ with either $\alpha = 1$ on the left panel or $\alpha=10$ on the right panel. The limiting spectra obtained using \eqref{eq:Stieltjes_hessian} are plotted as solid red lines and are perfectly fitting the two empirical distributions, together with their left-most edge characterized by the vertical dashed blue lines and obtained from \eqref{eq:left_edge}. The figure also depicts two regimes. In the left panel, the value of $\alpha$ is too small to observe an outlier outside of the bulk. In the right panel, an eigenvalue pops out of the continuous part of the Hessian spectrum, characteristic of the BBP transition that we analytically quantify in this appendix. This outlier eigenvalue is correctly predicted by \eqref{eq:Stieltjes_hessian_signal} as shown by the green cross in the figure. In Fig.~\ref{fig:hessian_spectrum_interm_times} we also show the bulks, left edges and outliers obtained for intermediate times while descending to threshold states during constrained initialization for $N=1024$ and normalization $a=0.01$. The left panel is obtained for $\alpha=3.6$ and $t=4$ where no detached eigenvalue exists, as predicted by the phase diagram in the right panel of Fig.~\ref{fig:overlaps}. By contrast, the right panel shows a situation where a clear outlier eigenvalue is detached from the bulk ($\alpha=7, \; t=8$), which is correctly predicted by our theory.

\subsection{Derivation of the overlap} \label{appendix:overlap}

To compute the squared overlap between the eigenvector associated to $\lambda_\star$, denoted $\bm{v_1}$, and the signal $\bm{w}^\star$ let us first remark that the problem is invariant by rotation. Hence we can focus only on the first component of the Stieltjes transform $\mathcal{S}_{11}(z)$ that can be decomposed using the eigenvectors $\{\bm{v}_i\}_{i=1}^N$ of $\mathcal{H}$ as
\begin{equation}
    \mathcal{S}_{11}(z) = \sum_{i=1}^N \frac{|\left[\bm{v}_i\right]_1 |^2}{z - \lambda_i}.
\end{equation}
which gives
\begin{equation}
    \lim_{z\rightarrow \lambda_\star} \mathcal{S}_{11}(z) = \frac{\left(\bm{v}_{1}\tran \bm{w}^\star\right)^2}{z-\lambda_\star}.
\end{equation}
By l'Hospital's rule,
\begin{equation}
    \left(\bm{v}_{1}\tran \bm{w}^\star\right)^2 = \lim_{z\rightarrow \lambda_\star} \frac{z-\lambda_\star}{ z - \Sigma(z)} = \frac{1}{1-\partial_z \Sigma(z)_{|z = \lambda_\star}},
\end{equation}
where $\Sigma(z)$ is given in \eqref{eq:Sigma_z}.

\section{Replica method for the computation of $p(y, \hat{y}, t_\mathrm{TS})$} \label{appendix:replicas}

In this Appendix, we aim to compute the probability distribution of the joint labels $p(y, \hat{y}, t_\mathrm{TS})$ on the threshold states that we conjecture to block the gradient flow dynamics in the large $N$ limit in Sect.~\ref{sect:RMT_BBP}. When $\alpha < \alpha_\mathrm{BBP}^{\mathrm{TS}}$, these states are defined as high-loss minima that are marginally stables (i.e., with a vanishing $\lambda_1$). To access this distribution, we rely on an heuristic method from statistical physics: \emph{the replica method}.
Let us first write the Boltzmann distribution associated to the system as
\begin{equation} \label{eq:boltzmann}
P(\bm{w}) = \frac{1}{Z(\beta)} \exp{-\beta \mathcal{L}(\bm{w})},
\end{equation}
where we denote $\bm{w}$ as shorthand notation for $\bm{w}^{(t)}$. $Z(\beta)$ is the partition function and $\mathcal{L}(\bm{w})$ is the \emph{energy} or \emph{cost function}.
The corresponding free energy per particle is
\begin{equation}
    \phi(\beta) = - \frac{1}{N\beta} \log Z(\beta),
\end{equation}
which is tightly coupled with many interesting macroscopic quantities of the system, like the average loss function, the expected overlap, but also to the joint probability distribution $p(y, \hat{y}, t)$ of true and estimated labels. As first explained in \cite{Franz2017} and also exploited in \cite{Mannelli2020b}, the typical distribution is given by $p(y, \hat{y}, t) = \mathbb{E}_{\bm{w}} \left[ \overline{\hat{p}(y, \hat{y}, t)} \right]$, where $\hat{p}$ denotes the empirical measure, the overline is the average over the disorder (here the dataset $\{\bm{x}_m\}_{m=1}^M$), and the expectation is taken over the Boltzmann measure. The partition function can be written in terms of $\hat{p}(y, \hat{y}, t)$ as
\begin{align}
    \overline{Z(\beta)} &= \overline{\int_{\mathbb{S}^{N-1}} \dd \bm{w} \exp{-\beta \mathcal{L}(\bm{w})}}, \\
    &= \overline{\int_{\mathbb{S}^{N-1}} \dd \bm{w} \exp{-\frac{\beta}{2} \sum_{m=1}^M \ell(y_i, \hat{y}_i)}}, \\
    &= \overline{\int_{\mathbb{S}^{N-1}} \dd \bm{w} \exp{-\frac{\beta M}{2} \int \dd y \dd \hat{y} \ell(y, \hat{y}) \hat{p}(y, \hat{y}, t) }}.
\end{align}

From this last expression, the distribution $p(y, \hat{y}, t)$ is accessible through the functional derivative of the free energy as
\begin{equation} \label{eq:free_energy_jpdf}
    \frac{\delta \overline{\phi}}{\delta \ell(y, \hat{y})} = -\frac{1}{N\beta} \frac{\delta\overline{\log Z(\beta)}}{\delta \ell(y, \hat{y})} = \frac{\alpha}{2} \mathbb{E}_{\bm{w}} \left[ \overline{\hat{p}(y, \hat{y}, t)} \right] = \frac{\alpha}{2} p(y, \hat{y}, t).
\end{equation}

This gives us some motivation for the computation of the log partition function, and more precisely its first moment if we can expect large deviation principle to apply to obtain the \emph{typical} behavior of the system.

\subsection{Replicated partition function}

To compute the average free energy per particle, we can use the replica method stating that
\begin{equation}
    \overline{\log Z} = \lim_{n\rightarrow 0} \frac{\overline{Z^n} - 1}{n}.
\end{equation}
In practice, we will compute $\overline{Z^n}$ for $n \in \mathbb{N}$ and then analytically continue it to $n \in \mathbb{R}$ in order to finally take the $n\rightarrow 0$ limit. The problem now boils down to compute $\overline{Z^n}$ which can be expressed as the partition function associated to the product of $n$ independent systems with the partition function $Z(\beta)$ and gives
\begin{equation}
    Z(\beta)^n = \int_{\mathbb{S}^{N-1}} \prod_{a=1}^n \left[ \dd \bm{w} \exp{-\beta \sum_{m=1}^M \ell\left(\bm{x}_m \cdot \bm{w}^{\star}, \bm{x}_m \cdot \bm{w}\right)} \right].
\end{equation}

Let us introduce $r_m^{(a)} = \bm{x}_i \cdot \bm{w}^{(a)}$, the overlap between the entries and the state of the $a$\textsuperscript{th} system, reserving the index zero for the overlap with the ground truth, meaning with $r_m^{(0)} = \bm{x}_i\tran \bm{w}^\star$. These new variables are introduced through delta functions that we replace by their Fourier representation. We therefore get
\begin{multline}
    Z(\beta)^n \propto \int_{\mathbb{S}^{N-1}} \prod_{a=1}^n \dd \bm{w}^{(a)} \int \prod_{a=0}^n \prod_{m=1}^M \dd r_m^{(a)} \int \prod_{a=0}^n \prod_{m=1}^M \dd \hat{r}_m^{(a)} \\
    \exp{-\beta \sum_{a=1}^n \sum_{m=1}^M \ell\left(r_m^{(0)}, r_m^{(a)}\right)
    + i \sum_{a=0}^n \sum_{m=1}^M \hat{r}_m^{(a)} r_m^{(a)} + i \sum_{a=0}^n \sum_{m=1}^M \hat{r}_m^{(a)} \bm{x}_i \cdot \bm{w}^{(a)}}.
\end{multline}
This allows us to compute the expectation over the disorder since, now, it only acts on the last term in the exponential. This integral can be evaluated using the Hubbard-Stratonovich identity \footnote{Stating that $\int \exp{-a x^2 + b x} \dd x = \sqrt{\pi/a} \exp{b^2/4a}$.} as
\begin{align}
    E(X) &= \mathbb{E}_{\bm{X}}\left[\exp{i \sum_{a=0}^n \sum_{m=1}^M \hat{r}_m^{(a)} \bm{x}_i \cdot \bm{w}^{(a)}}\right], \\
    &\propto \exp{- \frac{1}{2N} \sum_{a,b=0}^n \sum_{m=1}^M \hat{r}_m^{(a)} \hat{r}_m^{(b)} \bm{w}^{(a)} \cdot \bm{w}^{(b)}}.
\end{align}

Let us now consider the overlap between two replicas, $q_{ab} = \frac{1}{N} \bm{w}^{(a)} \cdot \bm{w}^{(b)}$. Similarly as previously, we use the index zero for the overlap with the signal $\bm{w}^\star$ such that $\forall a \in \left[1, n\right], q_{0a} = m$ and we also have $\forall a \in \left[0, n\right], q_{aa} = 1$. All these overlaps are regrouped into an $(n+1)\times (n+1)$ matrix $\bm{Q}$ and are introduced through a delta function again. It then reads
\begin{multline}
    \overline{Z(\beta)^n} \propto \int \prod_{0\leq a \leq b \leq n} \dd q_{ab} \int \prod_{a=0}^n \prod_{m=1}^M \dd r_m^{(a)} \int \prod_{a=0}^n \prod_{m=1}^M \dd \hat{r}_m^{(a)} \\ \underbrace{\int_{\mathbb{S}^{N-1}} \prod_{a=1}^n \dd \bm{w}^{(a)} \prod_{0\leq a \leq b \leq n} \delta(N q_{ab} - \bm{w}^{(a)} \cdot \bm{w}^{(b)})}_{J(\bm{Q})} \\
    \exp{-\beta \sum_{a=1}^n \sum_{m=1}^M \ell\left(r_m^{(0)}, r_m^{(a)}\right)
    + i \sum_{a=0}^n \sum_{m=1}^M \hat{r}_m^{(a)} r_m^{(a)} - \frac{1}{2} \sum_{a,b=0}^n \sum_{m=1}^M \hat{r}_m^{(a)} \hat{r}_m^{(b)} q_{ab}},
\end{multline}
with $J(\bm{Q}) = |\bm{Q}|^{N/2}$ in the large $N$ limit \cite{Zamponi2010}, consequently giving, after factorizing the $M$ integrals
\begin{multline} \label{eq:partition_generic}
    \overline{Z(\beta)^n} \propto \int \prod_{0\leq a \leq b \leq n} \dd q_{ab} \exp{\frac{N}{2}\log |\bm{Q}|} \Bigg[ \int \prod_{a=0}^n \dd r^{(a)} \int \prod_{a=0}^n \dd \hat{r}^{(a)} \\ 
    \exp{-\beta \sum_{a=1}^n \ell\left(r^{(0)}, r^{(a)}\right)
    + i \sum_{a=0}^n \hat{r}^{(a)} r^{(a)} - \frac{1}{2} \sum_{a,b=0}^n \hat{r}^{(a)} \hat{r}^{(b)} q_{ab}}\Bigg]^M.
\end{multline} 

Performing the integral over $\hat{r}$ using the Hubbard-Stratonovich identity again and setting $N\to\infty$, we finally obtain the replicated partition function
\begin{equation}  \label{eq:partition_function}
    \overline{Z(\beta)^n} \propto \exp{N\operatorname*{extr}_{\bm{Q}} S(\bm{Q})},
\end{equation}
with
\begin{equation}  \label{eq:action}
    S(\bm{Q}) = S_1 + S_2,
\end{equation}
and
\begin{align}
    S_1 &= \frac{1}{2} \log |\bm{Q}|, \\
    S_2 &= \alpha \log \int \prod_{a=0}^n \frac{\dd r^{(a)}}{\left(2\pi\right)^{n/2}\sqrt{|\bm{Q}|}}
    \exp{-\beta \sum_{a=1}^n \ell\left(r^{(0)}, r^{(a)}\right) - \frac{1}{2} \sum_{a,b} r^{(a)} \bm{Q}^{-1}_{ab} r^{(b)}}.
\end{align}
Here, $S_1$ is an entropic factor counting the number of spherical couplings that satisfies the constraints $q_{ab} = \bm{w}^{(a)} \cdot \bm{w}^{(a)} / N$ and $S_2$ is the energetic contribution specific to the learning rule in which appears the energy function per variable $\ell$. Notice that we turned the initial problem of computing a high-dimensional integral into a high-dimensional optimization over $(n+1)^2$ variables in \eqref{eq:action}. Although this may seem doomed, we can purse our analytical treatment by using an ansatz on the form of $\bm{Q}$. 

\subsection{One-step replica symmetry breaking (1RSB) ansatz}

The simplest form of hypothesis is called \emph{replica symmetry}, assuming $q_{ab} = q_0$ for $a\neq b$. However, this assumption breaks in the regime we are in and one needs to \emph{break the symmetry}. In our case, we use the first level of symmetry breaking (1RSB) assuming
\begin{equation}
\bm{Q} = \bm{Q}_{\mathrm{1RSB}} = 
  \renewcommand{\arraystretch}{1.2}
  \left(
  \begin{array}{ c | c c c c c c }
    1 & \multicolumn{2}{c}{m} & \multicolumn{2}{c}{\cdots} & \multicolumn{2}{c}{m} \\
    \cline{1-7}
     & \multicolumn{2}{c|}{\raisebox{.6\normalbaselineskip}[0pt][0pt]{}} & q_0 & q_0 & q_0 & q_0 \\
    \raisebox{.6\normalbaselineskip}[0pt][0pt]{$m$} & \multicolumn{2}{c|}{\raisebox{.6\normalbaselineskip}[0pt][0pt]{$\tilde{\bm{Q}}$}} & q_0 & q_0 & q_0 & q_0 \\
    \cline{2-5}
    
    & q_0 & q_0 & \multicolumn{1}{|c}{} & \multicolumn{1}{c|}{} & q_0 & q_0 \\
    \raisebox{.6\normalbaselineskip}[0pt][0pt]{$\vdots$} & q_0 & q_0 & \multicolumn{2}{|c|}{\raisebox{.6\normalbaselineskip}[0pt][0pt]{$\tilde{\bm{Q}}$}} & q_0 & q_0 \\
    \cline{4-7}
    
    & q_0 & q_0 & q_0 & q_0 &  \multicolumn{1}{|c}{} & \multicolumn{1}{c|}{} \\
    \raisebox{.6\normalbaselineskip}[0pt][0pt]{$m$} & q_0 & q_0 & q_0 & q_0 & \multicolumn{2}{|c|}{\raisebox{.6\normalbaselineskip}[0pt][0pt]{$\tilde{\bm{Q}}$}} \\
    \cline{6-7}
  \end{array}
  \hspace{0.05cm} \right) \in \mathbb{R}^{(n+1)\times(n+1)},
\end{equation}
with $\tilde{\bm{Q}}$ a matrix of size $p \times p$ with one on the diagonal and $q_1$ everywhere else. Under this assumption, the action can be written in terms of the four parameters $n$, $m$, $p$, $q_0$ and $q_1$. This hence reduces the saddle point method to extremize over those parameters only in \ref{eq:partition_function}. This type of matrix was extensively studied in statistical physics, and one result of particular interest for us is that $\bm{Q}_{\mathrm{1RSB}}$ has three eigenvalues $\tilde{\lambda}_i$ with multiplicities $d_i$ given by \cite{Castellani2005}
\begin{equation}
    \begin{cases}
    \tilde{\lambda}_1 = 1 - q_1, & d_1 = n\left(1-\frac{1}{p}\right),\\
    
    \tilde{\lambda}_2 = p(q_1 - q_0) + (1-q_1), & d_2 = \frac{n}{p} - 1,\\
    
    \tilde{\lambda}_3 = p(q_1 - q_0) + (1-q_1) + n(q_0-m^2), & d_3 = 1.\\
    
    \end{cases}\,
\end{equation}
Using these eigenvalues, we can evaluate the entropy in the action as
\begin{multline} \label{eq:S1_1RSB}
S_1^{\mathrm{1RSB}}(q_0, q_1, m, p) = \frac{n}{2} \Bigg[ \log(1-q_1) + \frac{1}{p} \log \frac{1-q_1 + p(q_1-q_0)}{1-q_1} \\ + \frac{q_0-m^2}{1-q_1 + p(q_1-q_0)} \Bigg] + O(n^2).
\end{multline}
For the energetic term, one has to use the form of $\bm{Q}_\mathrm{1RSB}$ to work out that
\begin{multline}
    -\frac{1}{2} \sum_{a,b} \hat{r}^{(a)} \hat{r}^{(b)} q_{ab} = -\frac{1}{2} \Bigg[ \hat{r}^{(0)}\hat{r}^{(0)} + (1-q_1) \sum_{a=1}^n \hat{r}^{(a)}\hat{r}^{(a)} \\ + (q_1 - q_0) \sum_{P_a=1}^{n/p} \left( \sum_{a\in P_a} \hat{r}^{(a)} \right)^2 + q_0 \left(\sum_{a=1}^n \hat{r}^{(a)}\right)^2 + 2m \sum_{a=1}^n r^{(a)}r^{(0)} \Bigg].
\end{multline}
Substuting it into $S_2$ gives
\begin{multline} \label{eq:S2_1RSB}
    S_2^{\mathrm{1RSB}}(q_0, q_1, m, p) \underset{n\rightarrow 0^{+}}{\approx} \frac{\alpha n}{p} \int \dd \eta \int \dd r^{(0)} D(r^{(0)}, \eta) \log\Bigg(\int \frac{\dd \eta_P} {\sqrt{2\pi(q_1-q_0)}} \\ \exp{-\frac{\eta_P^2}{2(q_1-q_0)}}
    \left[ \int \frac{\dd r}{\sqrt{2\pi}} \exp{-\beta \Psi(r^{(0)}, r, \eta_P, \eta, q_1)} \right]^p \Bigg),
\end{multline}
where
\begin{equation}
    \Psi(r^{(0)}, r, \eta_P, \eta, q_1) = \ell(r^{(0)}, r) + \frac{(\eta_P + \eta - r)^2}{2\beta(1-q_1)}.
\end{equation}

\subsection{Zero-temperature limit and free energy}

The 1RSB free energy is defined as the zero temperature limit ($\beta\to\infty$) of the extremum of the 1RSB action $S^\mathrm{1RSB} = S(\bm{Q}_\mathrm{1RSB})$
\begin{align}
    \phi_{\mathrm{1RSB}} &= \lim_{\beta\rightarrow+\infty} \lim_{n\rightarrow 0^{+}} -\frac{1}{n\beta}  S_1^{\mathrm{1RSB}}(q_0, q_1, m, p) -\frac{1}{n\beta} S_2^{\mathrm{1RSB}}(q_0, q_1, m, p).
\end{align}
While taking the $\beta \rightarrow +\infty$ limit, we set $q_1 \rightarrow 1$ keeping both $\chi=\beta(1-q_1)$ and $z=\beta p$ of order one. Putting it all together, and setting $m$ to zero by remarking it satisfies the saddle-point $\partial_m S^{\mathrm{1RSB}} = 0$, we end up with the 1RSB free energy
\begin{multline}
    \phi_{\mathrm{1RSB}}(\chi, z, q_0) = -\frac{1}{2z} \log\frac{\chi + z(1-q_0)}{\chi} - \frac{1}{2} \frac{q_0}{\chi + z(1-q_0)} - \frac{\alpha}{z} \int \dd \eta \int \dd r^{(0)} D(r^{(0)}, \eta) \\
    \log\left( \int \frac{\dd \eta_P}{\sqrt{2\pi (1-q_0)}} \exp{-\frac{\eta_P^2}{2(1-q_0)} - z \Psi_0(r^{(0)}, \eta_P, \eta, \chi)}\right),
\end{multline}
with
\begin{align}
    \Psi_0(r^{(0)}, \eta_P, \eta, \chi) &= \min_{\tilde{r}} \ell(r^{(0)}, \tilde{r}) + \frac{(\eta_P + \eta - \tilde{r})^2}{2\chi}, \\
    D(r^{(0)}, \eta) &= \frac{1}{2\pi\sqrt{q_0}} \exp{-\frac{{r^{(0)}}^2 q_0\eta + \eta^2}{2q_0}}.
\end{align}

From \eqref{eq:free_energy_jpdf}, we need to take the functional derivative of the free energy with respect to the loss function $\ell(y, \hat{y})$ to obtain the joint distribution of true and estimated labels on threshold states $p(y, \hat{y}, t_\mathrm{TS})$.
This gives
\begin{multline} \label{eq:P_1RSB}
    p(y, \hat{y}, t_\mathrm{TS}) = \frac{1}{\sqrt{2\pi}} \int \frac{\dd \eta}{\sqrt{2\pi q_0}} \exp{-\frac{y^2 q_0 + \eta^2}{2q_0}} \\ \frac{\exp{-\frac{\hat{y}^2}{2(1-q_0)} - z\Psi_0(y, \hat{y}, \eta, \chi)}}{\int \dd \tilde{y} \exp{-\frac{\tilde{y}^2}{2(1-q_0)} - z\Psi_0(y, \tilde{y}, \eta, \chi)}},
\end{multline}
which is equivalent to the finding of \cite{Mannelli2020b} if we set $q_0 = 0$.
Finally, the parameters $\chi$, $z$, and $q_0$ are fixed via the saddle-point equations obtained from $\partial_{\chi} S^\mathrm{1RSB} = 0$ and $\partial_{q_0} S^\mathrm{1RSB} = 0$, giving
\begin{multline}
    \frac{1}{z} \left(\frac{1}{\chi} - \frac{1}{\chi+z(1-q_0))}\right) + \frac{q_0}{(\chi + z(1-q_0))^2} = \alpha \int \dd \eta \int \dd r^{(0)} D(r^{(0)}, \eta) \\
    \frac{\int \dd \eta_P \exp{-\frac{\eta_P^2}{2(1-q_0)} - z\Psi_0(r^{(0)}, \eta_P, \eta, \chi)} \left(\partial_{\eta_P}\Psi_0(r^{(0)}, \eta_P, \eta, \chi)\right)^2}{\int \dd \eta_P \exp{-\frac{\eta_P^2}{2(1-q_0)} - z\Psi_0(r^{(0)}, \eta_P, \eta, \chi)}},
\end{multline}
\begin{multline}
    -\frac{q_0}{(\chi+z(1-q_0))^2} = \frac{2\alpha}{z^2} \int \dd \eta \int \dd r^{(0)} D(r^{(0)}, \eta) \frac{\eta^2 - q_0}{2q_0^2}
    \log \Bigg( \int \frac{\dd \eta_P}{\sqrt{2\pi (1-q_0)}} \\ \exp{-\frac{\eta_P^2}{2(1-q_0)} - z\Psi_0(r^{(0)}, \eta_P, \eta, \chi)} \Bigg) + \frac{\alpha}{z^2(1-q_0)}
    - \frac{\alpha}{z^2(1-q_0)^2} \\ \times \int \dd \eta \int \dd r^{(0)} D(r^{(0)}, \eta) \frac{\int \dd \eta_P \exp{-\frac{\eta_P^2}{2(1-q_0)} - z\Psi_0(r^{(0)}, \eta_P, \eta, \chi)} \eta_P^2 }{\int \dd \eta_P \exp{-\frac{\eta_P^2}{2(1-q_0)} - z\Psi_0(r^{(0)}, \eta_P, \eta, \chi)}}.
\end{multline}
Finally, to actually probe the threshold states instead of the global minima of the landscape when taking the $\beta\to\infty$ limit, one has to fix the parameter $z$ using the marginal stability condition of the Hessian, as first shown in \cite{cugliandolo1993analytical} and used in \cite{Franz2017, Mannelli2020b}. This grants access to the probability distribution for $p(y, \hat{y}, t=t_\mathrm{TS})$. Using \eqref{eq:P_1RSB} in equations~\eqref{eq:BBP_generic_1} and~\eqref{eq:BBP_generic_2} yields the value $\alpha_\mathrm{BBP}^\mathrm{1RSB, TS} = 4.29$. We expect that breaking further the symmetry by assuming substructures in $\bm{Q}_\mathrm{1RSB}$ would reduce the gap with the $\alpha_\mathrm{BBP}^\mathrm{TS} = 4.03$ obtained from the sampling of threshold states but leave this aspect for further investigations.

\section{Details of the numerical experiments} \label{appendix:numerical}

\begin{figure}
        \centering
        \includegraphics[width=.49\linewidth]{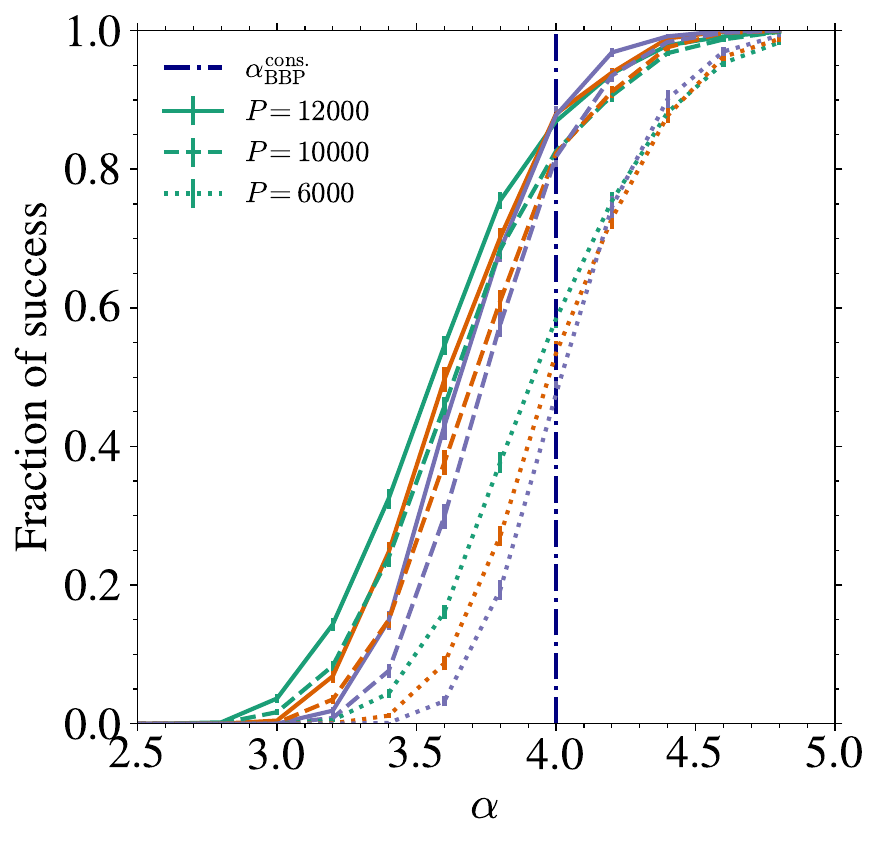}
        \caption{Fraction of successes as a function of $\alpha$ for different simulation times $T=P\log_2(N)$ in the constrained initialization. Green corresponds to $N=512$ orange to $N=1024$, and purple to $N=2048$. The vertical dashed line shows $\alpha_\mathrm{cons.}^\mathrm{SR} \approx 4.0$ used in the main text for $P=12,000$.}
        \label{fig:varying_T}
\end{figure}

All the numerical experiments were run on CPUs for $N\leq1024$ while on NVIDIA A6000 for $N\geq2048$. Depending on the value of $N$, $\alpha$, and on the initialization scheme, it takes between fifteen minutes to sixteen hours to obtain a batch of $100$ simulations.
All the models are trained using gradient descent with fixed learning rate $\eta$ and a total number of $T$ steps, starting from an initial condition $\hat{\bm{w}}^{(0)}$ that takes three different forms: random, constrained, or spectral. For random initialization, $\hat{\bm{w}}^{(0)} \sim \mathcal{N}(0, \bm{I}_N)$ while for spectral initialization $\hat{\bm{w}}^{(0)} = \bm{v}_1$, the eigenvector associated to the smallest eigenvalue of the Hessian from a random state. Finally, for the constrained initialization, we use Algorithm~\ref{alg:constrained_init} with $t_\mathrm{c}=60,000$ steps allowing to reach a threshold state $\hat{w}^{(t_\mathrm{c})}$ that we use as initial condition for standard gradient descent. The number of steps after initialization is $T=P\log_2(N)$ where $P=12,000$ in the main text. It is varied from $6,000$ to $12,000$ in Fig.~\ref{fig:varying_T} showing the convergence of fraction of successes when $P$ increases. In practice, we find that when $P>9,000$ the transition is always found at $\alpha_\mathrm{cons.}^\mathrm{SR}\approx4.0$ (shown in vertical dashed line).

\begin{algorithm}
    \caption{Constrained initialization}\label{alg:constrained_init}
    \begin{algorithmic}
        \Require $\alpha = M/N > 0$, $t_\mathrm{c}>0$, $\eta>0$
        \State $\bm{w}^{(0)} \gets \mathcal{N}(0,\bm{I}_N)$
        \State $t\gets 0$
        \While{$t < t_\mathrm{c}$}
        \State $\nabla\mathcal{L}(\bm{w}^{(t)}) \gets \frac{1}{2}\sum_{i=1}^M \nabla_{\hat{\bm{w}}^{(t)}}\ell(y_i, \hat{y}_i)$ 
        \State $\mu^{(t)} \gets \bm{w}^{(t)} \cdot \nabla\mathcal{L}(\bm{w}^{(t)}) / N$
        \State $\bm{w}^{(t+1)} \gets \bm{w}^{(t)} - \eta\nabla\mathcal{L}(\hat{\bm{w}}^{(t)}) + \eta \mu^{(t)}\hat{\bm{w}}^{(t)}$
        \EndWhile
    \end{algorithmic}
\end{algorithm}

\begin{figure}
        \centering
        \includegraphics[width=.49\linewidth]{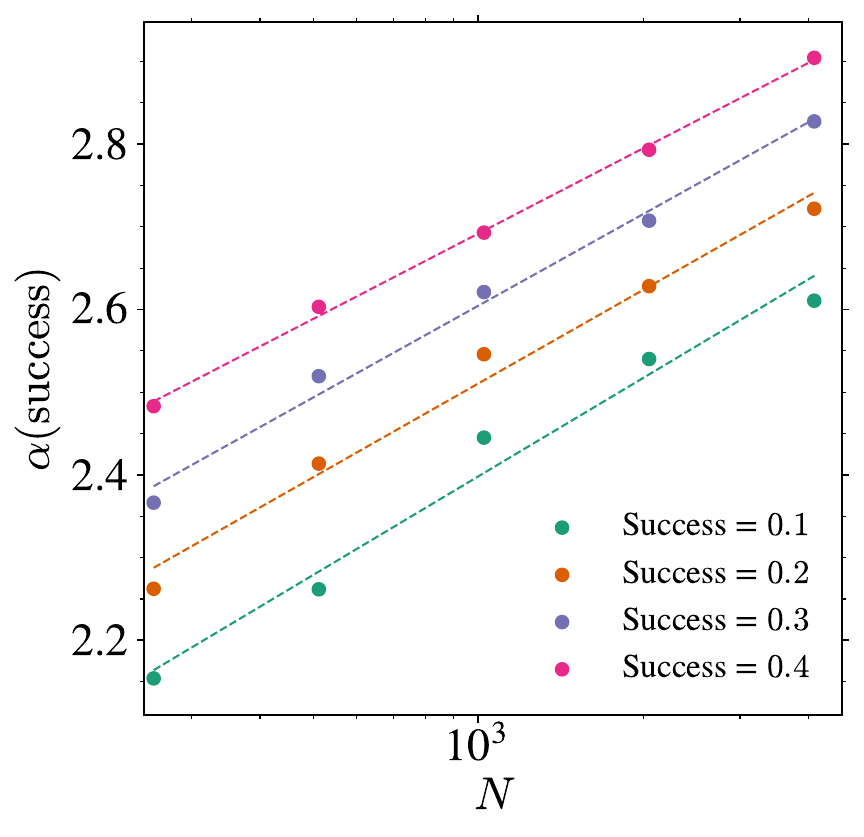}
        \includegraphics[width=.49\linewidth]{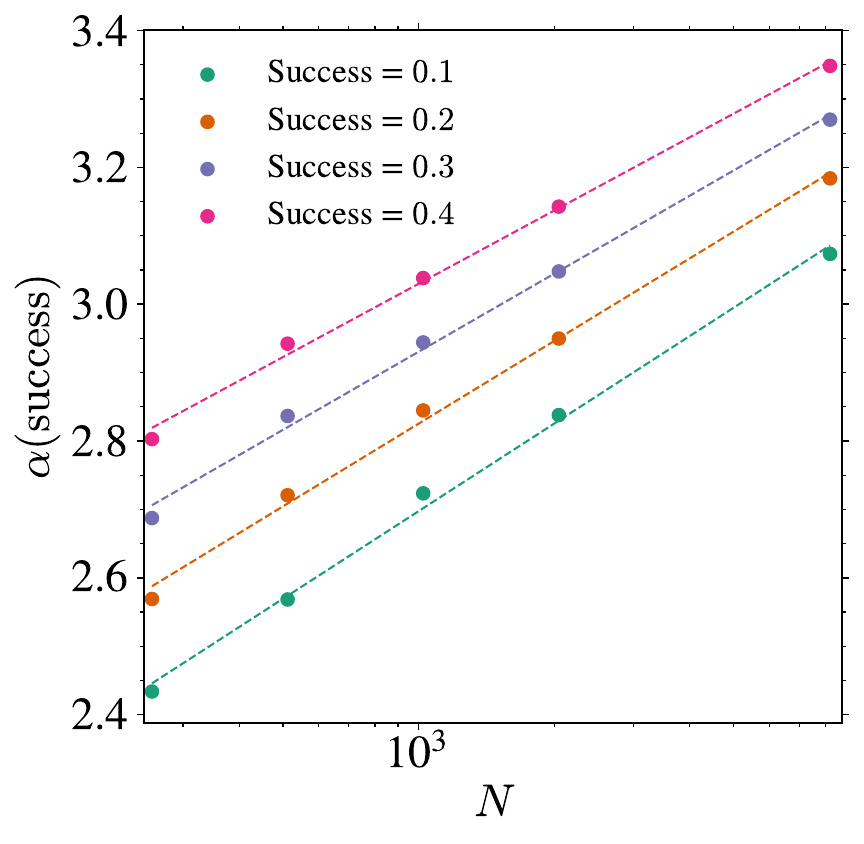}
        \vspace{-.4cm}
        \caption{Scaling of $\alpha$ for several fixed strong recovery rates (shown in Figure~\ref{fig:fraction_success}) for random initializations and $N \in \{256, 512, 1024, 2048, 4096, 8192\}$ for \emph{(Left)} $a=0.01$ and \emph{(Right)} $a=1$.}
        \label{fig:Log_scale_transition}
\end{figure}

\paragraph*{Logarithmic scaling of the strong recovery rates.}
In Figure~\ref{fig:Log_scale_transition} can be found some evidence of the displacement of the strong recovery rates obtained in Figure~\ref{fig:fraction_success} for randomly initialized weights with $N \in \left[256, 8192\right]$. In this case, the effective transition is shown to scale as $\log N$ for two very different values of $a$ (0.01 as in the main text, and 1), as a consequence of the local initial curvature coupled with the initial magnetization $m(0)$ of order $1/\sqrt{N}$, as discussed at the end of Section~\ref{sect:RMT_BBP}.

\paragraph*{Numerical estimate of the BBP transition on threshold states.}
In Section~\ref{sect:RMT_BBP} and Section~\ref{sect:Numerical}, we use a numerical approach to extract $p(y, \hat{y}, t_\mathrm{TS})$ and compute $\alpha_{\mathrm{BBP}}^\mathrm{TS}$. The method relies on sampling the threshold states using the constrained initialization (see Section~\ref{subsect:constrained}) to then compute the expectations from equations~\eqref{eq:BBP_generic_1},~\eqref{eq:BBP_generic_2}, and ~\eqref{eq:overlap} by averaging numerically. Of course, this means that we are using finite $N$ simulations to compute expectations derived for $N\to\infty$. In practice, we use $N=\{512, 1024, 2048\}$ simulations to perform a finite-size scaling analysis of $\alpha_\mathrm{BBP}^{(t)}$. We checked that this procedure allows us to retrieve the analytical value of $\alpha_\mathrm{BBP}^\mathrm{init} = 2.85$ with great accuracy and obtain on threshold states the value given in the main text of $\alpha_\mathrm{BBP}^\mathrm{TS}=4.03$. In order to check the consistency with larger values of $N$, we also compared this result with hundreds of numerical simulations with $N=8192$ leading to the same value.

\section{Weak recovery in spectral initialization} \label{appendix:spectral}
\begin{figure}
    \centering
    \includegraphics[width=.49\linewidth]{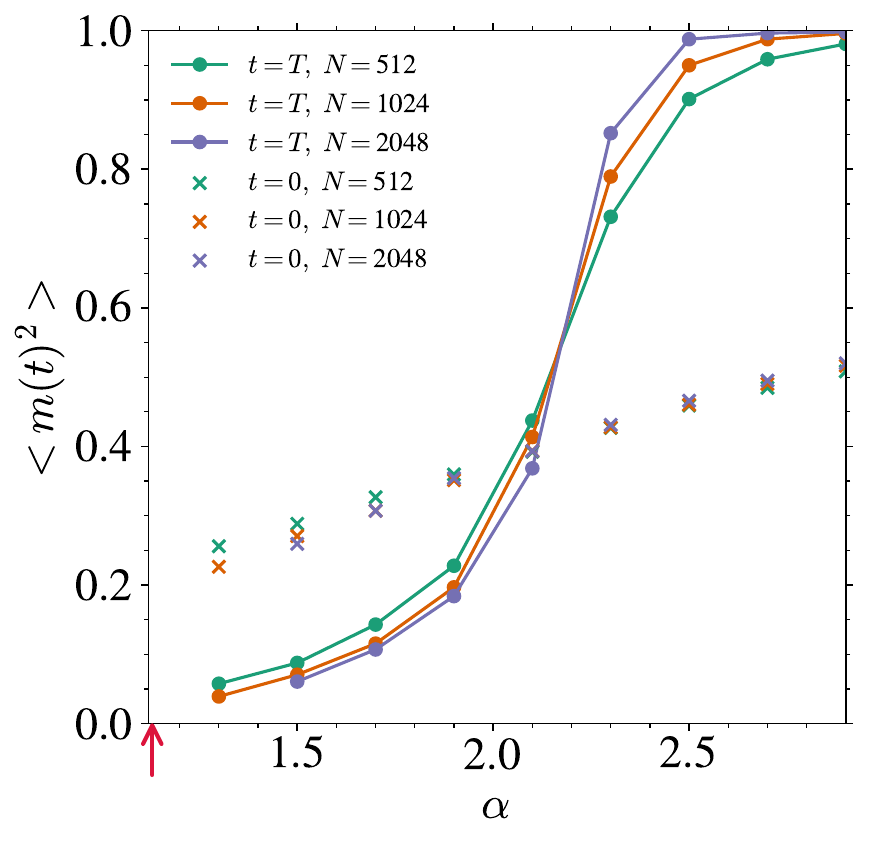}
    \includegraphics[width=.49\linewidth]{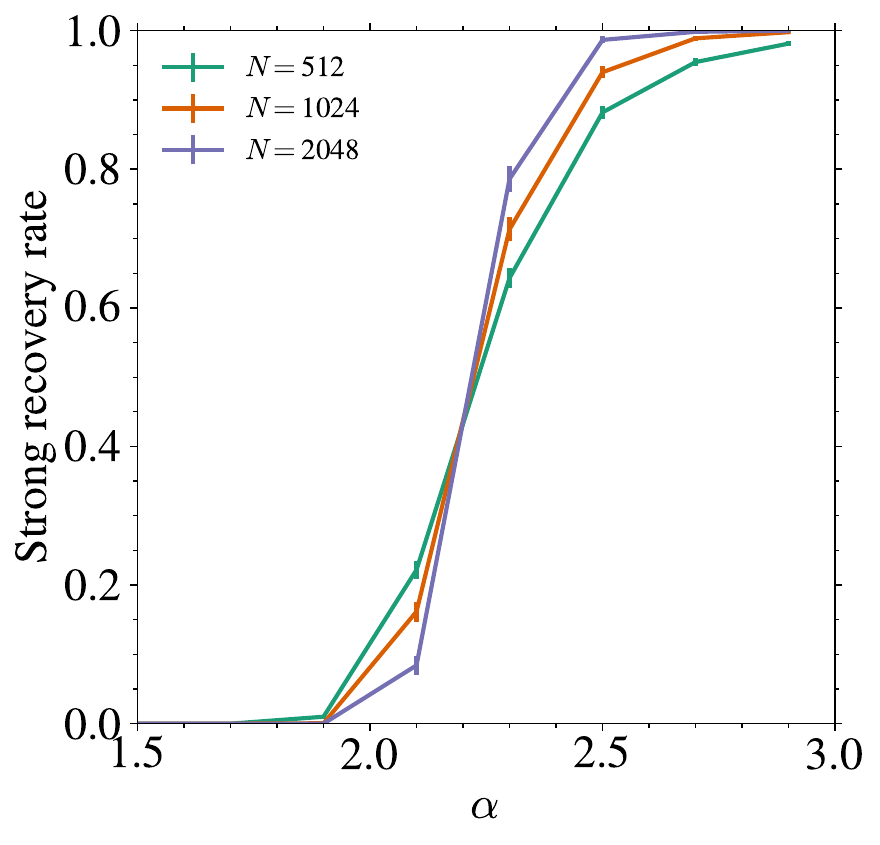}
    \vspace{-.4cm}
    \caption{\emph{(Left)} Averaged squared magnetization $\langle m(t)^2 \rangle$ as a function of $\alpha$ for several values of $N$ at times $0$ and $T$ using spectral initialization along $\bm{v}_1^{(0)}$. \emph{(Right)} Strong recovery rates for spectral initialization with different values of $N$. Both plots are obtained with $a=1$ in the normalization of the loss \eqref{eq:loss}.
    }
    \label{fig:Spectral_mag_a=1}
\end{figure}

In Section~\ref{sect:Numerical}, we highlight the importance of a good initial guess to efficiently solve the phase retrieval problem, and we advocate for the existence of an intermediary phase where the estimate performs weak recovery. While this effect is not obvious in the main text because the strong recovery rate starts to increase roughly at the same $\alpha$ as $\langle m(T)^2 \rangle$ in Figure~\ref{fig:Spectral_mag}, it is however clearer for larger $a$, as illustrated in Figure~\ref{fig:Spectral_mag_a=1}. When $\alpha$ is small (around $2$), and no success is yet observed, the magnetization $\langle m(t)^2 \rangle$ already takes significant values of around 0.3. For $\alpha < 2$, we also observe an interesting phenomenon where the initial guess has more overlap with the signal than at the end of the gradient descent dynamics, also suggesting a rough landscape outside the equator for such SNRs.

\section{Impact of the loss function on the BBP transitions} \label{appendix:loss}

\begin{figure}
    \centering
    \includegraphics[width=.49\linewidth]{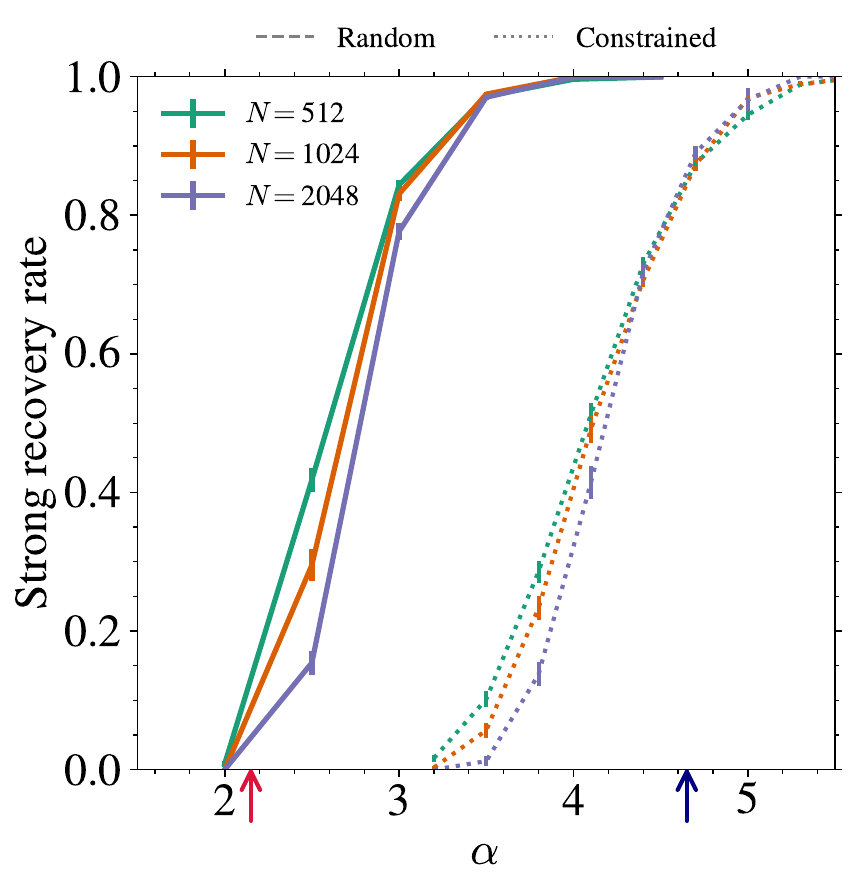}
    \includegraphics[width=.49\linewidth]{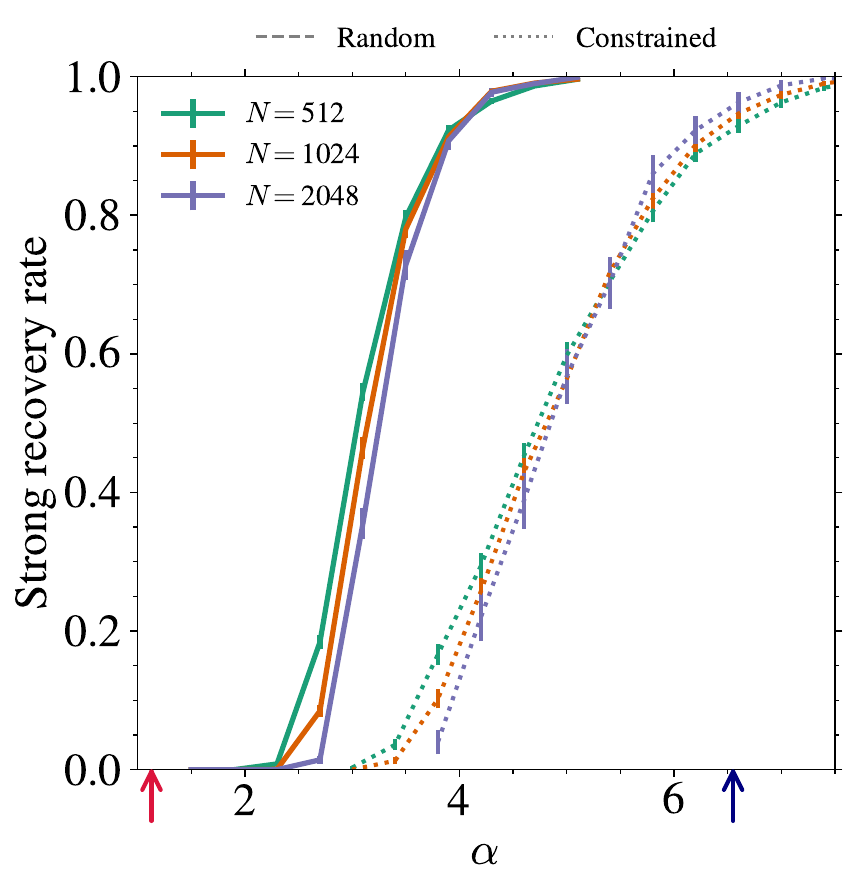}
    \vspace{-.4cm}
    \caption{Strong recovery rates for the loss function $\ell_a$ defined in \eqref{eq:loss} with \textit{(Left)} $a=0.1$ and \textit{(Right)} $a=1$ for random (solid lines) and constrained (dotted lines) initializations. The red (resp. blue) arrow indicates $\alpha_\mathrm{BBP}^\mathrm{init}$ (resp. $\alpha_\mathrm{BBP}^\mathrm{TS}$) computed in these cases.
    }
    \label{fig:BBP_loss}
\end{figure}

\begin{figure}
    \centering
    \includegraphics[width=.49\linewidth]{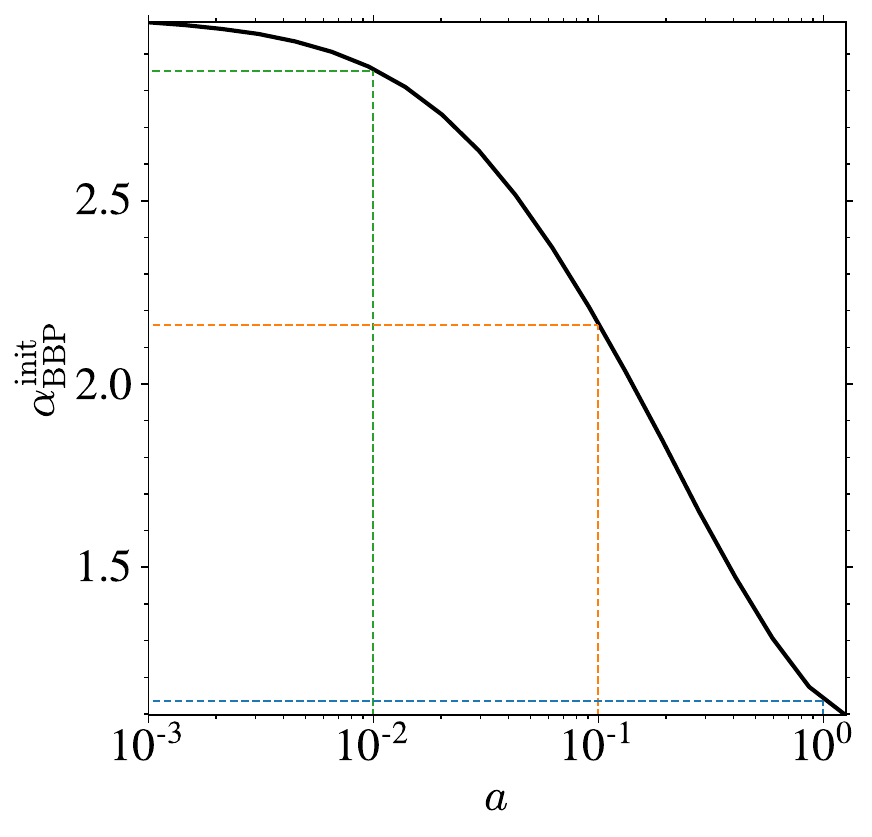}
    \vspace{-.4cm}
    \caption{Evolution of the BBP threshold transition at initialization $\alpha_\mathrm{BBP}^\mathrm{init}$ for various values of $a$ in the loss function. The three values discussed in the main text and the appendix are displayed as colored dashed lines for $a=0.01$, $a=0.1$, and $a=1$.
    }
    \label{fig:BBP_a_init}
\end{figure}

In the main text, we focused on the loss function $\ell_a(y, \hat{y})$ from \eqref{eq:loss} with $a=0.01$. The precise values of the BBP transitions at both initialization and on threshold states however depend on the second derivative of $\ell$ and some choices may lead to more favorable landscapes enabling earlier strong recovery. To illustrate this, we plot in Figure~\ref{fig:BBP_loss} the strong recovery rates obtained with several values of $a$ for the loss function \eqref{eq:loss}. In particular, increasing $a$ from $0.01$ in the main text to $0.1$ or $1$ (respectively left and right panels) leads to lower $\alpha_\mathrm{BBP}^\mathrm{init}$, meaning less samples are required to start having the local curvature towards the signal at initialization. For $a=0.1$, we find $\alpha_\mathrm{BBP}^\mathrm{init} = 2.16$ while $\alpha_\mathrm{BBP}^\mathrm{init} = 1.13$ for $a=1$. More values of $\alpha_\mathrm{BBP}^\mathrm{init}$ are shown as a function of $a$ in Fig.~\ref{fig:BBP_a_init}. Larger $a$ allows more favorable landscape at initialization by decreasing the required SNR to observe the first BBP transition at $t=0$.

Even though the initial states have a downward direction towards the signal at lower values of the signal-to-noise ratio, threshold states on their side develop an instability later for increasing $a$. In particular, we find $\alpha_\mathrm{BBP}^\mathrm{TS} = 4.03$ for $a=0.01$ (main text scenario), $\alpha_\mathrm{BBP}^\mathrm{TS} = 4.65$ for $a=0.1$ and $\alpha_\mathrm{BBP}^\mathrm{TS} = 6.55$ for $a=1$. This is also clearly seen in the constrained simulations of Figure~\ref{fig:BBP_loss} where the algorithmic transition occurs later than in Figure~\ref{fig:fraction_success} for both random and constrained initializations. In these cases, we also observe a logarithmic scaling of success rates with $N$ for random initializations while the successes are delayed with the constrained initialization and the curves for different $N$ intersect nicely. Finally, we note that the predicted value of $\alpha_\mathrm{BBP}^\mathrm{TS}$ obtained from \eqref{eq:BBP_generic_1} using the numerical simulations to compute the expectations is matching less precisely the algorithmic threshold of the constrained simulations than in the main text. This is particularly true for $a=1$ where $\alpha_\mathrm{cons.}^\mathrm{SR} \approx 5.55$, inducing a gap with the $N\to\infty$ prediction.

\bibliographystyle{iopart-num}
\bibliography{library}

\providecommand{\newblock}{}
\begin{thebibliography}{10}
\expandafter\ifx\csname url\endcsname\relax
  \def\url#1{{\tt #1}}\fi
\expandafter\ifx\csname urlprefix\endcsname\relax\def\urlprefix{URL }\fi
\providecommand{\eprint}[2][]{\url{#2}}

\bibitem{Fyodorov2004}
Fyodorov Y~V 2004 {\em Physical Review Letters\/} {\bf 93}(14) 149901--149901 ISSN 0031-9007

\bibitem{Rico2007}
Rico F and Moy V~T 2007 {\em Journal of Molecular Recognition\/} {\bf 20} 495--501

\bibitem{Auffinger2010}
Auffinger A, Arous G~B and Cerny J 2010  (\textit{Preprint} \eprint{1003.1129})

\bibitem{Baity-Jesi2019}
Baity-Jesi M, Sagun L, Geiger M, Spigler S, Arous G~B, Cammarota C, LeCun Y, Wyart M and Biroli G 2019 {\em Journal of Statistical Mechanics: Theory and Experiment\/} {\bf 12}(12) 124013 ISSN 0201-7563

\bibitem{Gardner1988}
Gardner E and Derrida B 1988 {\em Journal of Physics A: General Physics\/} {\bf 21}(1) 271--284 ISSN 0305-4470

\bibitem{Seung1992}
Seung H~S, Sompolinsky H and Tishby N 1992 {\em Phys. Rev. A\/} {\bf 45}(8) 6056--6091 \urlprefix\url{https://link.aps.org/doi/10.1103/PhysRevA.45.6056}

\bibitem{Krzakala2009_hiding}
Krzakala F and Zdeborov\'a L 2009 {\em Phys. Rev. Lett.\/} {\bf 102}(23) 238701 \urlprefix\url{https://link.aps.org/doi/10.1103/PhysRevLett.102.238701}

\bibitem{Zdeborova2018}
Zdeborova L and Krzakala F 2016 {\em Advances in Physics\/} {\bf 65} 453--552

\bibitem{Neyshabur2017}
Neyshabur B, Bhojanapalli S, McAllester D and Srebro N 2017 Exploring generalization in deep learning {\em Proceedings of the 31st International Conference on Neural Information Processing Systems\/} NIPS'17 (Red Hook, NY, USA: Curran Associates Inc.) pp 5949--5958 ISBN 9781510860964

\bibitem{Belkin2018}
Belkin M, Ma S and Mandal S 2018 To understand deep learning we need to understand kernel learning {\em Proceedings of the 35th International Conference on Machine Learning\/} ({\em Proceedings of Machine Learning Research\/} vol~80) ed Dy J and Krause A (PMLR) pp 541--549

\bibitem{ma2018power}
Ma S, Bassily R and Belkin M 2018 The power of interpolation: Understanding the effectiveness of sgd in modern over-parametrized learning {\em International Conference on Machine Learning\/} (PMLR) pp 3325--3334

\bibitem{Venturi2019}
Venturi L, Bandeira A~S and Bruna J 2019 {\em Journal of Machine Learning Research\/} {\bf 20} 1--34

\bibitem{Mannelli2020a}
Mannelli S~S, Vanden-Eijnden E and Zdeborová L 2020 {\em Advances in Neural Information Processing Systems\/} {\bf 2020-Decem} 1--26 ISSN 1049-5258

\bibitem{Martin2024}
Martin S, Bach F and Biroli G 2024 On the impact of overparameterization on the training of a shallow neural network in high dimensions {\em Proceedings of The 27th International Conference on Artificial Intelligence and Statistics\/} ({\em Proceedings of Machine Learning Research\/} vol 238) ed Dasgupta S, Mandt S and Li Y (PMLR) pp 3655--3663 \urlprefix\url{https://proceedings.mlr.press/v238/martin24a.html}

\bibitem{Annesi2024}
Annesi B~L, Lauditi C, Lucibello C, Malatesta E~M, Perugini G, Pittorino F and Saglietti L 2023 {\em Phys. Rev. Lett.\/} {\bf 131}(22) 227301

\bibitem{Soudry2016}
Soudry D and Carmon Y 2016 {\em arXiv preprint arXiv:1605.08361\/}

\bibitem{Cai2022}
Cai J, Huang M, Li D and Wang Y 2022 {\em Applied and Computational Harmonic Analysis\/} {\bf 58} 60--84 (\textit{Preprint} \eprint{2101.03540})

\bibitem{liu2020bad}
Liu S, Papailiopoulos D and Achlioptas D 2020 Bad global minima exist and sgd can reach them {\em Advances in Neural Information Processing Systems\/} vol~33 ed Larochelle H, Ranzato M, Hadsell R, Balcan M and Lin H (Curran Associates, Inc.) pp 8543--8552

\bibitem{Ros2019}
Ros V, Arous G~B, Biroli G and Cammarota C 2019 {\em Physical Review X\/} {\bf 9}(1) 11003 ISSN 2160-3308

\bibitem{Mannelli2019}
Mannelli S~S, Biroli G, Cammarota C, Krzakala F and Zdeborová L 2019 {\em Advances in Neural Information Processing Systems\/} {\bf 32} 1--28 ISSN 1049-5258

\bibitem{Mannelli2020b}
Mannelli S~S, Biroli G, Cammarota C, Krzakala F, Urbani P and Zdeborová L 2020 {\em Advances in Neural Information Processing Systems\/}  1--17 ISSN 1049-5258

\bibitem{Baik2005}
Baik J, Arous G~B and Péché S 2005 {\em Annals of Probability\/} {\bf 33}(5) 1643--1697 ISSN 0091-1798

\bibitem{Millane1990}
Millane R~P 1990 {\em Journal of the Optical Society of America Part A\/} {\bf 7}(3) 394--411

\bibitem{Harrison1993}
Harrison R~W 1993 {\em Journal of the Optical Society of America Part A\/} {\bf 10}(5) 1046--1055

\bibitem{Miao2008}
Miao J, Ishikawa T, Shen Q and Earnest T 2008 {\em Annual Review of Physical Chemistry\/} {\bf 59}(November 2007) 387--410 ISSN 0066-426X

\bibitem{Shechtman2014}
Shechtman Y, Eldar Y~C, Cohen O, Chapman H~N, Miao J and Segev M 2014 {\em arXiv e-prints\/}  1--25 (\textit{Preprint} \eprint{1402.7350})

\bibitem{Fienup2019}
Fienup J~R 2019 Phase retrieval for image reconstruction {\em Imaging and Applied Optics 2019 (COSI, IS, MATH, pcAOP)\/} (Optica Publishing Group) p CM1A.1

\bibitem{Wong2021}
Wong A, Pope B, Desdoigts L, Tuthill P, Norris B and Betters C 2021 {\em Journal of the Optical Society of America B\/} {\bf 38}(9) 2465 ISSN 0740-3224

\bibitem{Pardalos1991}
Pardalos P~M and Vavasis S~A 1991 {\em Journal of Global Optimization\/} {\bf 1}(1) 15--22 ISSN 0925-5001

\bibitem{Candes2015}
Candès E~J, Li X and Soltanolkotabi M 2015 {\em IEEE Transactions on Information Theory\/} {\bf 61}(4) 1985--2007 ISSN 0018-9448

\bibitem{Netrapalli2015}
Netrapalli P, Jain P and Sanghavi S 2015 {\em IEEE Transactions on Signal Processing\/} {\bf 63}(18) 4814--4826 ISSN 1053-587X (\textit{Preprint} \eprint{1306.0160})

\bibitem{Waldspurger2015}
Waldspurger I, D’Aspremont A and Mallat S 2015 {\em Mathematical Programming\/} {\bf 149}(1-2) 47--81 ISSN 1436-4646

\bibitem{Chen2017a}
Chen Y and Candès E~J 2017 {\em Communications on Pure and Applied Mathematics\/} {\bf 70}(5) 822--883 ISSN 1097-0312

\bibitem{Zhang2017}
Zhang H, Zhou Y, Liang Y and Chi Y 2017 {\em Journal of Machine Learning Research\/} {\bf 18} 1--35 ISSN 1533-7928

\bibitem{Wang2017}
Wang G, Giannakis G~B and Chen J 2017 {\em 25th European Signal Processing Conference, EUSIPCO 2017\/} {\bf 2017-Janua}(1) 1420--1424

\bibitem{Wang2017a}
Wang G, Giannakis G~B, Saad Y and Chen J 2017 {\em Advances in Neural Information Processing Systems\/} {\bf 2017-Decem} 1868--1878 ISSN 1049-5258

\bibitem{Zhang2018}
Zhang C, Wang M, Chen Q, Wang D and Wei S 2018 {\em International Journal of Optics\/} {\bf 2018} ISSN 1687-9392

\bibitem{BenArous2021}
Arous G~B, Gheissari R and Jagannath A 2021 {\em Journal of Machine Learning Research\/} {\bf 22} 1--51

\bibitem{BenArous2022}
{Ben Arous} G, Gheissari R and Jagannath A 2022 High-dimensional limit theorems for {SGD}: Effective dynamics and critical scaling {\em Advances in Neural Information Processing Systems\/} ed Oh A~H, Agarwal A, Belgrave D and Cho K

\bibitem{Bietti2022}
Bietti A, Bruna J, Sanford C and Song M~J 2022 Learning single-index models with shallow neural networks {\em Advances in Neural Information Processing Systems\/} ed Oh A~H, Agarwal A, Belgrave D and Cho K

\bibitem{Arnaboldi2023}
Arnaboldi L, Krzakala F, Loureiro B and Stephan L 2023 {\em arXiv preprint arXiv:2305.18502\/} (\textit{Preprint} \eprint{2305.18502})

\bibitem{Bruna2023}
Bruna J, Pillaud-Vivien L and Zweig A 2023 On single index models beyond gaussian data (\textit{Preprint} \eprint{2307.15804})

\bibitem{Barbier2019}
Barbier J, Krzakala F, Macris N, Miolane L and Zdeborová L 2019 {\em Proceedings of the National Academy of Sciences of the United States of America\/} {\bf 116}(12) 5451--5460 ISSN 1091-6490

\bibitem{Mondelli2019}
Mondelli M and Montanari A 2019 {\em Foundations of Computational Mathematics\/} {\bf 19}(3) 703--773 ISSN 1615-3383

\bibitem{Luo2018}
Luo W, Alghamdi W and Lu Y~M 2019 {\em IEEE Transactions on Signal Processing\/} {\bf 67}(9) 2347--2356 (\textit{Preprint} \eprint{1811.04420})

\bibitem{maillard2020phase}
Maillard A, Loureiro B, Krzakala F and Zdeborov{\'a} L 2020 {\em Advances in Neural Information Processing Systems\/} {\bf 33} 11071--11082

\bibitem{maillard2022construction}
Maillard A, Krzakala F, Lu Y~M and Zdeborov{\'a} L 2022 Construction of optimal spectral methods in phase retrieval {\em Mathematical and Scientific Machine Learning\/} (PMLR) pp 693--720

\bibitem{Sun2018}
Sun J, Qu Q and Wright J 2018 {\em Foundations of Computational Mathematics\/} {\bf 18}(5) 1131--1198 ISSN 1615-3383

\bibitem{Li2020}
Li Z, Cai J~F and Wei K 2020 {\em IEEE Transactions on Information Theory\/} {\bf 66}(5) 3242--3260 ISSN 1557-9654

\bibitem{Cai2021}
Cai J~F, Huang M, Li D and Wang Y 2021 {\em arXiv e-prints\/} (1) 1--41 (\textit{Preprint} \eprint{2112.07997})

\bibitem{Saade2014}
Saade A, Krzakala F and Zdeborová L 2014 {\em Advances in Neural Information Processing Systems\/} {\bf 27}(January) 406--414

\bibitem{Bun2017}
Bun J, Bouchaud J~P and Potters M 2017 {\em Physics Reports\/} {\bf 666} 1--109 ISSN 0370-1573

\bibitem{SaraoMannelli2020}
Mannelli S~S, Biroli G, Cammarota C, Krzakala F, Urbani P and Zdeborová L 2020 {\em Physical Review X\/} {\bf 10}(1) 1--45 ISSN 2160-3308

\bibitem{fraboul2023artificial}
Fraboul J, Biroli G and De~Monte S 2023 {\em Journal of Theoretical Biology\/} {\bf 571} 111557

\bibitem{bonnaire2023highdimensional}
Bonnaire T, Ghio D, Krishnamurthy K, Mignacco F, Yamamura A and Biroli G 2023 High-dimensional non-convex landscapes and gradient descent dynamics (\textit{Preprint} \eprint{2308.03754})

\bibitem{Lu2020}
Lu Y~M and Li G 2020 {\em Information and Inference: A Journal of the IMA\/} {\bf 9}(3) 507--541 ISSN 2049-8772

\bibitem{Mannelli2019a}
Mannelli S~S, Krzakala F, Urbani P and Zdeborova L 2019 Passed and spurious: Descent algorithms and local minima in spiked matrix-tensor models {\em Proceedings of the 36th International Conference on Machine Learning\/} ({\em Proceedings of Machine Learning Research\/} vol~97) ed Chaudhuri K and Salakhutdinov R (PMLR) pp 4333--4342

\bibitem{Peche2006}
Péché S 2006 {\em Journal of Multivariate Analysis\/} {\bf 97}(4) 874--894 ISSN 0047-259X

\bibitem{Franz2017}
Franz S, Parisi G, Sevelev M, Urbani P and Zamponi F 2017 {\em SciPost Physics\/} {\bf 2}(3) 1--37 ISSN 2542-4653

\bibitem{Luo2020}
Luo Q, Lin S and Wang H 2021 {\em Symmetry\/} {\bf 13}(11) ISSN 2073-8994

\bibitem{Mignacco2021}
Mignacco F, Urbani P and Zdeborová L 2021 {\em Machine Learning: Science and Technology\/} {\bf 2}(3) ISSN 2632-2153

\bibitem{Ghorbani2019}
Ghorbani B, Krishnan S and Xiao Y 2019 An investigation into neural net optimization via hessian eigenvalue density {\em Proceedings of the 36th International Conference on Machine Learning\/} ({\em Proceedings of Machine Learning Research\/} vol~97) ed Chaudhuri K and Salakhutdinov R (PMLR) pp 2232--2241

\bibitem{Sun2020optimization}
Sun R~Y 2020 {\em Journal of the Operations Research Society of China\/} {\bf 8} 249--294

\bibitem{Adahessian2021}
Yao Z, Gholami A, Shen S, Mustafa M, Keutzer K and Mahoney M 2021 {\em Proceedings of the AAAI Conference on Artificial Intelligence\/} {\bf 35} 10665--10673

\bibitem{Maillard2019b}
Maillard A, Ben~Arous G and Biroli G 2020 Landscape complexity for the empirical risk of generalized linear models {\em Proceedings of The First Mathematical and Scientific Machine Learning Conference\/} ({\em Proceedings of Machine Learning Research\/} vol 107) ed Lu J and Ward R (PMLR) pp 287--327

\bibitem{Cai2023}
Cai J~F, Huang M, Li D and Wang Y 2023 {\em IOP Publishing\/} {\bf 39}(7) 075011

\bibitem{BenArous2025_local}
Arous G~B, Gheissari R, Huang J and Jagannath A 2025 Local geometry of high-dimensional mixture models: Effective spectral theory and dynamical transitions (\textit{Preprint} \eprint{2502.15655})

\bibitem{Zamponi2010}
Zamponi F 2010 {\em arXiv e-prints\/} (\textit{Preprint} \eprint{1008.4844})

\bibitem{Castellani2005}
Castellani T and Cavagna A 2005 {\em Journal of Statistical Mechanics: Theory and Experiment\/} (5) 215--266 ISSN 1742-5468

\bibitem{cugliandolo1993analytical}
Cugliandolo L~F and Kurchan J 1993 {\em Physical Review Letters\/} {\bf 71} 173

\end{thebibliography}

\end{document}